%% file: main.tex
\documentclass{article}

\PassOptionsToPackage{numbers, compress}{natbib}

\usepackage[preprint]{neurips_2022}

\usepackage[utf8]{inputenc} 
\usepackage[T1]{fontenc}    
\usepackage{hyperref}       
\usepackage{url}            
\usepackage{booktabs}       
\usepackage{amsfonts}       
\usepackage{nicefrac}       
\usepackage{microtype}      
\usepackage{xcolor}         

\input{tikzstuff.tex}
\input{math_commands.tex}

\usepackage{amsmath,bm}

\usepackage{hyperref}
\usepackage{graphicx}
\usepackage{subfigure}
\usepackage{float}
\usepackage{dsfont}

\usepackage{makecell}
\usepackage{multirow}

\usepackage{pgfplots}
\usepgfplotslibrary{fillbetween}
\usetikzlibrary{patterns}

\usepackage{hyperref}

\title{Representation Learning for High-Dimensional Data Collection under Local Differential Privacy}

\begin{document}


\author{%
\textbf{Alex Mansbridge}$^{1,2}$\thanks{Correspondence to: \texttt{amansbridge@turing.ac.uk}} \quad \textbf{Davide Piras}$^{3}$\thanks{Equal Contribution} \quad \textbf{Gregory Barbour}$^{3 \dagger}$ \quad \textbf{Michael Murray}$^4$ \\
\textbf{Christopher Frye}$^3$ \quad \textbf{Ilya Feige}$^3$ \quad \textbf{David Barber}$^1$ \\
$^1$Department of Computer Science, University College London \hspace{2mm} $^2$Alan Turing Institute, London, UK
\\
$^3$Faculty, 160 Old Street, London, UK \hspace{2mm} $^4$Department of Mathematics, UCLA
}





\maketitle

\begin{abstract}
The collection of individuals’ data has become commonplace in many industries. Local differential privacy (LDP) offers a rigorous approach to preserving privacy whereby the individual privatises their data locally, allowing only their perturbed datum to leave their possession. LDP thus provides a provable privacy guarantee to the individual against both adversaries and database administrators. Existing LDP mechanisms have successfully been applied to low-dimensional data, but in high dimensions the privacy-inducing noise largely destroys the utility of the data. In this work, our contributions are two-fold: first, by adapting state-of-the-art techniques from representation learning, we introduce a novel approach to \emph{learning} LDP mechanisms. These mechanisms add noise to powerful representations on the low-dimensional manifold underlying the data, thereby overcoming the prohibitive noise requirements of LDP in high dimensions. Second, we introduce a novel denoising approach for downstream model learning. The training of performant machine learning models using collected LDP data is a common goal for data collectors, and downstream model performance forms a proxy for the LDP data utility. Our approach significantly outperforms current state-of-the-art LDP mechanisms.
\end{abstract}

\section{Introduction}
\label{Section:Intro}
The collection of personal data is ubiquitous, and unavoidable for many in everyday life. The use of such data for training machine learning algorithms has become instrumental in improving the quality and user experience of many products and services. However, evidence of data misuse and data breaches \citep{Jolly2020, Sweeney1997} has brought the concept of data privacy into sharp focus, fuelling regulatory changes, and a shift in consumer preferences. There is thus a growing need for data collection methods that preserve both individuals' privacy and data utility for product and service improvement.

Privatising data under \emph{local differential privacy} (LDP) \citep{Duchi2013, Kasiviswanathan2008} naturally lends itself to data collection. LDP mechanisms allow individuals to privatise their data before sharing it, thus providing a mathematically-provable privacy guarantee for the individual against both a potential adversary and the database administrator. LDP has its roots in randomised response \citep{Warner1965}, which preserves the privacy of survey respondents by only having them answer a sensitive binary question truthfully if a secret coin flip returns heads. 
Limited research has gone into developing LDP mechanisms for high-dimensional data, especially those that generalise to different data types. Often dubbed the ``curse of dimensionality'', this is a challenging problem in LDP \citep{Bhowmick2019, Duchi2018, Zhang2017}. 

The local Laplace mechanism \cite{Dwork2006} is the de-facto standard for continuous attributes. While \citet{Duchi2018} and \citet{Wang2019} introduce lower variance continuous LDP mechanisms, \citet{Duchi2018} emphasise the pessimistic nature of results in high dimensions, and \citet{Wang2019}'s mechanism entails collecting $k \ll d$ perturbed attributes per $d$-dimensional datapoint (where for experiments in this paper $k\leq4$, but $d$ can be over 3000). The $\text{PrivUnit}_2$ mechanism \cite{Bhowmick2019} privatises high-dimensional continuous gradient data for federated learning; the authors consider only very high local-$\epsilon$ guarantees, aiming to protect only against accurate data reconstruction rather than arbitrary inferences. 

There exist a number of LDP mechanisms for specific tasks or data types. \citet{Ding2017} study the repeated collection of one-dimensional telemetry data for histogram estimation. \citet{Erlingsson2014} collect aggregate statistics on categorical attributes, with \citet{Fanti2016} extending this to model correlations between dimensions, but neither produce representations suitable for downstream learning on high-dimensional data. \citet{Ren2018} discuss the poor performance of \cite{Erlingsson2014, Fanti2016} in high dimensions, and instead estimate the distribution of collected data, from which they sample a synthetic dataset. The range of applications here is limited (see next paragraph), and the approach incurs a high communication cost between the data collector and individuals. In summary, developing a general method for inducing LDP in high dimensions, while preserving utility, is an open question.

\emph{Central} differential privacy (CDP) \citep{Dwork2006} is  a  related  framework offering  protection  in  an  altogether  different context.  Rather than facilitating the private collection of individual datapoints, CDP mechanisms stop an adversary determining, up to a quantifiable level of certainty, the presence of an individual in a dataset. This is achieved via the calibrated addition of noise to the output of queries on that dataset. However, to achieve this requires the database administrator have access to the full unprivatised dataset. CDP has been used effectively in the related field of data \emph{release}. For example, \citet{Xie2018, Triastcyn2018, Acs2019, Takagi2021} propose releasing synthetic data composed of samples from generative models trained with a CDP optimisation algorithm \citep{Abadi2016, Gylberth2017, Papernot2017}. While powerful in some scenarios, this approach is not suited to data collection, where we are trying to protect individuals from all external parties, \emph{including} the database administrator. Furthermore, the synthetic data provides no information about the features of specific individuals, and the distribution of the synthetic dataset is static after training the generative model.

In this paper, we introduce a suite of entirely novel, \emph{learnt} LDP mechanisms that adapt techniques from representation learning. We motivate our approach with two observations: first, it is often a good approximation to assume high-dimensional data lies on a low-dimensional manifold. Second, the vast majority of organisations collecting personal data already have access to \emph{auxiliary data} that could be used to learn the low-dimensional manifold underlying the data distribution. This may be data scraped from the internet, as is commonly used to train unsupervised models \cite{Devlin2019, Mahajan2018, Ramesh2021}, public datasets \cite{Deng2009, Irvin2019, Thomee2016}, or previously collected internal data \cite{CMA2020, Schmidt2018}. We introduce a novel de-noising approach for downstream model training on our privatised data (a common goal of data collectors), and use downstream model performance as a proxy to measure the utility of our LDP training data.

Our approach adapts state-of-the-art representation learning techniques to map high-dimensional data to representations on a constrained, low-dimensional manifold, before adding LDP-inducing noise. Mapping to such a manifold in this way has three key advantages: First, we circumvent the pessimistic nature of results for provably optimal LDP mechanisms in high-dimensions \cite{Bhowmick2019, Duchi2018}. Second, we can train our mechanism to ensure the representations learnt are maximally robust to LDP-inducing noise. Finally, in practice, these state-of-the-art methods learn extremely powerful representations when compared to traditional dimensionality-reduction techniques. In the absence of noise, downstream models trained using such representations have even been shown to match performance of models trained with supervised learning \cite{Chen2020}. 

Our approach provides a clear, simple framework for adapting a multitude of existing representation learning techniques. Consequently, owing to a wealth of existing representation learning research, our mechanism can be applied to a broad range of data types, including images \citep{Chen2020, He2020, Kingma2014}, audio \citep{Vandenoord2017, Vandenoord2018}, text \citep{Bowman2016, Giorgi2021}, and video \citep{Denton2018}. Finally, the generalisation ability of such models 
(a) reduces the extent to which auxiliary data must follow the same distribution as the data being collected and (b) allows the collected LDP data to be used for an array of downstream tasks. In particular, we demonstrate our approach learns powerful LDP data representations, significantly outperforming state-of-the-art LDP benchmark mechanisms on three major applications:
\begin{itemize}
    \vspace{-2.0mm}
    \item Data privatised with our mechanisms is used to train downstream machine learning models, demonstrating state-of-the-art utility preservation on high dimensional data under LDP.
    \vspace{-1.0mm}
    \item Our mechanisms can be used, without re-training, to collect distributionally-shifted data. In particular, we train classifiers on privatised data for novel class classification.
    \vspace{-1.0mm}
    \item As a mechanism for LDP, it can track the IDs of real individuals whilst privatising their sensitive features. We use this to augment internal data with privatised features from an external source to improve a classifier's performance on the combined feature set.
\end{itemize}

\begin{figure*}[t]
  \centering
  \includegraphics[width=\linewidth]{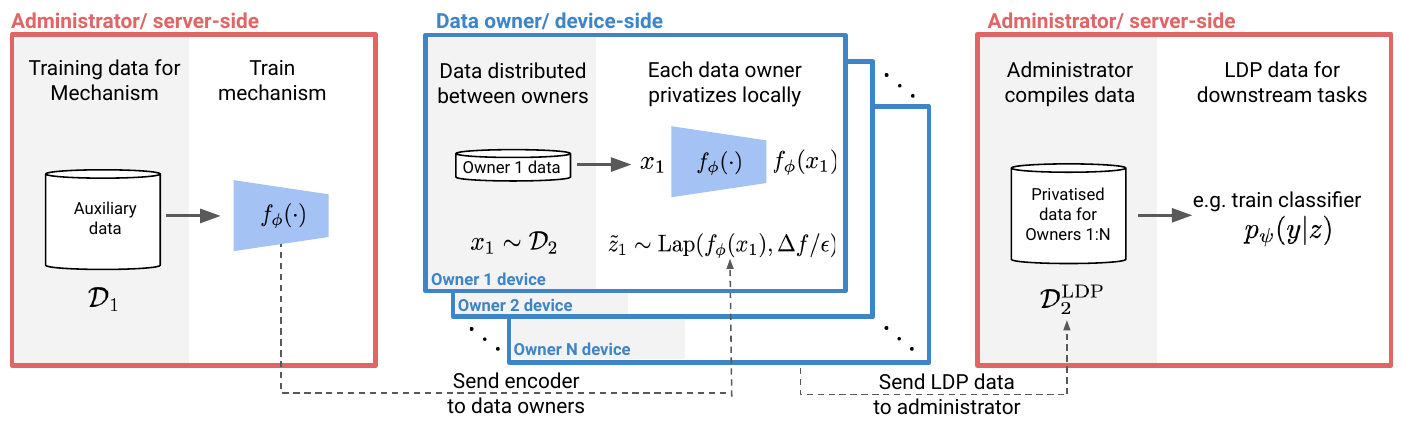}
  \vspace{-8mm}
  \caption{Schematic diagram of mechanism training (left), local data privatisation (centre) and collection (right), as outlined in Section \ref{Section:method}. Red boxes indicate operations performed on the administrator/ data collector's infrastructure and blue boxes indicates operations performed locally by the data owner. Crucially, unprivatised data never leaves the data owner's device.}
  \label{fig:vlm_diagram}
\end{figure*}

\section{Definitions and Notation}
\label{Section:definitions}
To formalise the concept of local differential privacy, we first introduce some definitions and notation.




\textbf{Definition ($\epsilon$-local differential privacy):} A local randomised algorithm $\mathcal{A}: \mathcal{X} \rightarrow \mathcal{Z}$,  that takes as input a datapoint from the data domain $\mathcal{X}$, satisfies $\epsilon$-local differential privacy if for $\epsilon \geq 0$, for any $S \subseteq \mathcal{Z}$, and for all inputs $x, x' \in \mathcal{X}$, 
\begin{equation}
    p(\mathcal{A}(x) \in S) \leq e^{\epsilon} \, p(\mathcal{A}(x') \in S) \label{eqn:LDP}
\end{equation}
Intuitively, this states that one cannot tell (with a level of certainty determined by $\epsilon$) whether the output of a local randomised algorithm $\mathcal{A}$ is the privatised version of a datapoint $x \in \mathcal{X}$, or the privatised version of any other input $x' \in \mathcal{X}$.

\textbf{Definition (Local $\ell_1$ sensitivity):} The $\ell_1$ sensitivity of a function $f: \mathcal{X} \rightarrow \mathcal{Z}$, where $\mathcal{Z} \subseteq \mathbb{R}^k$, is defined as
\begin{equation}
    \Delta f = \max_{x, x' \in \mathcal{X}} || f(x) - f(x')||_1
\end{equation}
\textbf{Definition (Local Laplace mechanism):} The local Laplace mechanism $\mathcal{M}: \mathcal{X} \rightarrow \mathbb{R}^k$ is a randomised algorithm defined as
\begin{equation}
   \mathcal{M}\left(x, f(\cdot), \epsilon \right) = f(x) + (s_1, \hdots, s_k) \ \label{eqn:local_laplace_mechanism}
\end{equation}
for $x \in \mathcal{X}$, $s_i \sim \text{Laplace}(0, \Delta f/\epsilon)$, and some transformation function $f: \mathcal{X} \rightarrow \mathcal{Z}$ with local $\ell_1$ sensitivity $\Delta f$. The local Laplace mechanism satisfies $\epsilon$-LDP (see Appendix \ref{appendix:laplace_mech_proof} for proof). 


\section{Proposed Method}
\label{Section:method}

Many existing LDP mechanisms noise each feature independently; by the composition theorem \citep{Dwork2014}, the $i^\text{th}$ feature contributes towards the overall LDP guarantee of the $d$-dimensional datapoint as $\epsilon = \sum_{i=1}^d \epsilon_i$. For fixed $\epsilon$, as $d$ increases, $\epsilon_i$ decreases for each feature $i$ and so the noise required to induce $\epsilon$-LDP grows. High-dimensional datapoints like images or large tables often contain highly correlated features; consequently, noising features independently is wasteful towards privatising the information content. Even if one does not noise features independently, data utility decreases as dimensionality increases, resulting in poor performance in high dimensions \cite{Bhowmick2019, Duchi2018}. Instead, we propose a more effective approach: learning an application-agnostic mechanism that privatises a range of data types through the addition of complex, non-linear noise on a constrained, low-dimensional manifold.

Individuals apply this learnt LDP mechanism to their data before sharing with the data collector, who then forms a LDP dataset composed of data collected from all individuals. This setup is agnostic to choice of downstream-task so can be applied broadly. We consider training downstream machine learning algorithms as the evaluative task for our LDP datapoints. Specifically, we use the LDP data, along with information on the type of noise added, and prior beliefs regarding the distribution of our representation space, to train a classifier network that takes features $x$ as input and predicts the corresponding label $y$. This classifier can be trained to act on either clean or privatised features at inference time, depending on the application.

\subsection{Learning a Laplace Mechanism}
\label{Section:rep_learning}
We introduce a framework for learning the mapping $f(\cdot)$ in Equation \ref{eqn:local_laplace_mechanism}, by adapting techniques from representation learning. In addition to learning task-agnostic data representations for downstream model training, the representations learnt must be robust to LDP-inducing Laplace noise. Equivalently, we must learn a mapping that maximally-separates (in L1-norm) representations of points with different features, such as images from different classes. 

We parameterise the mapping $f_{\phi}(\cdot):\mathcal{X} \rightarrow \mathcal{Z}$ from data space to a representation space using a neural network with parameters $\phi$. The sensitivity $\Delta f_{\phi}$ is controlled via a carefully chosen activation function $\nu(\cdot)$ on the final layer $f_{\phi}(\cdot) = \nu(h_\phi(\cdot))$:
\begin{equation}
    \nu(h) = \frac{l\, h}{||h||_1} \mathbb{I}\big[ ||h||_1 > l \big] + h \mathbb{I}\big[ ||h||_1 \leq l \big]
\end{equation}
Here, $\mathcal{Z}$ is the taxicab hypersphere containing all points within L1-distance $l$ from the origin, and $\Delta f_{\phi} = 2l$. For any $x \in \mathcal{X}$, adding $\text{Laplace}(0, 2l\mathbb{I} / \epsilon_x)$ noise to $z=f_\phi(x)$ is equivalent to passing $x$ through the local Laplace mechanism $\mathcal{M}\left(x, f_{\phi}(\cdot), \epsilon_x \right)$, guaranteeing $\epsilon_x$-LDP. We denote this LDP representation $\tilde{z}$, and dub this mechanism the Representation-Learning Laplace Mechanism (RLM).

Other mechanisms could be used by changing the activation $\nu(\cdot)$, and the notion of separation (i.e. noise robustness) in $\mathcal{Z}$. We use the Laplace mechanism since it is well-studied in the LDP literature and has stronger guarantees than, say, the Gaussian mechanism which satisfies only $(\epsilon, \delta>0)$-LDP (see e.g. \citet{Wang2021}). We leave the investigation of alternative mechanisms to future work.

 We introduce two approaches for learning $f_\phi(\cdot)$. The approach should be chosen by the data collector based on the data being collected: we focus on one approach suited to lower-dimensional datasets such as tabular data and small images, and another for higher-dimensional colour images. Other representation learning approaches can be easily adapted with our framework, allowing LDP collection for a plethora of datatypes. We emphasise that even tabular data is considered high dimensional for LDP mechanisms, owing to the aforementioned curse of dimensionality. 

In Sections \ref{sec:VLM}, we adapt a latent variable model (LVM) \cite{Kingma2014, Rezende2014} to induce LDP. LVMs have shown strong performance for representation learning on tabular data and small images. For higher dimensional data like colour images however, LVMs typically utilise deep hierarchies of latents, with the total number of latent dimensions often exceeding the original data dimension \cite{Child2021, Maaloe2019}. While these models achieve state-of-the-art performance in terms of log likelihood, they are not well suited to learning low-dimensional representations. To learn a LDP mechanism for higher-dimensional colour images, we instead use a contrastive approach. Such models are naturally suited here, having achieved state-of-the-art performance for representation learning on high-dimensional images \citep{Chen2020,He2020, Vandenoord2018}.

Training $f_\phi(\cdot)$ requires auxiliary data $\mathcal{D}_1$ with similar distribution to the data $\mathcal{D}_2$ we hope to collect. It is reasonable to assume most organisations can access such data, as discussed in Section \ref{Section:Intro}. In some instances, the data may be sensitive. Since we share $f_\phi(\cdot)$ with data owners, we must protect the privacy of these dataset members under CDP. In Appendix \ref{appendix:cdp_data_collection}, we give a formal definition of CDP, and demonstrate training $f_\phi(\cdot)$ under this privacy guarantee.

\subsubsection{Variational RLM (V-RLM)}
\label{sec:VLM}

We first introduce a variational approach to learning a Laplace mechanism. We assume each datapoint $x$ is generated by a random process involving a latent variable $z$ of dimension $d$. We then optimise a lower bound on the log likelihood \citep{Kingma2014, Rezende2014}
\begin{align}
    \log p(x) &\geq \mathbb{E}_{q_\phi(z|x)}\left[\log p_\theta(x|z) \right] -\KL\big(q_\phi(z|x)||p(z)\big)
\end{align}
where $p(z)$ is the prior distribution and $q_\phi(z|x)$ is the approximate posterior over the latent representation. The generative distribution $p_\theta(x|z)$ and approximate inference distribution $q_\phi(z|x)$ are parameterised by neural networks, with learnable parameters $\theta$ and $\phi$ respectively. Since we aim to learn a local Laplace mechanism, we choose
\begin{align}
    p(z) = \prod_{i=1}^d p(z_i)
    \quad \text{and} \quad
    q_\phi(z|x) = \prod_{i=1}^d q_\phi(z_i|x)
\end{align}
where $p(z_i) = \text{Laplace}(0,1/\sqrt{2})$ and $q_\phi(z_i|x) = \text{Laplace}(f_{\phi}(x)_i,b)$, where $f_{\phi}(\cdot)_i = \nu(h_\phi(\cdot))_i$.

Fixing $b=\Delta f_\phi/\epsilon_x=2l/\epsilon_x$ at privatisation time guarantees a sample from $q_\phi(z|x)$, which we denote $\tilde{z}$, is $\epsilon_x$-LDP. We experiment with different values of $b$ during mechanism training. Experiments conducted in Appendix \ref{appendix:PCA_experiments}, in which we privatise data using other dimensionality reduction techniques, strongly suggest that this addition of noise during training is important to learning a latent representation that is robust to the $\epsilon_x$-LDP noise requirements at privatisation time. We refer to this method as the Variational RLM (V-RLM).

We prove in Appendix \ref{appendix:LDP_post_processing} that a reconstruction $\tilde{x}$, obtained by passing $\tilde{z}$ through the decoder network $p_\theta(\cdot|z)$, also satisfies $\epsilon_x$-LDP. This allows the collection of privatised datapoints on either representation level $\tilde{z}$, or original-feature level $\tilde{x}$, depending on the data collector's preference. 

\subsubsection{Contrastive RLM (C-RLM)}
\label{sec:CLM}

Next, in order to learn noise-robust LDP representations for colour images, we adapt SimCLR \cite{Chen2020}. Again, we want to learn a mapping $f_{\phi}(\cdot) = \nu(h_\phi(\cdot))$ where $\nu(\cdot)$ constrains the output so that $\Delta f_{\phi} = 2l$, and additive $\text{Laplace}(0,2l\mathbb{I}/\epsilon_x)$ noise on $z =f_{\phi}(x)$ guarantees $\epsilon_x$-LDP.

Each image in the minibatch $\{x_b\}_{b=1}^{B}$, is augmented twice via randomly sampled transforms $t\sim \mathcal{T}$ (for details see Appendix \ref{apendix:transformations}). These augmentations are mapped through $f_{\phi}(\cdot)$ giving $2B$ representations $\{z_{b'}\}_{b'=1}^{2B}$, where for each positive pair (representations of augmentations of the same image) we have $2(B-1)$ negative pairs (representations of augmentations of different images).

As in the V-RLM, we add $\text{Laplace}(0,b\mathbb{I})$ noise to these representations during mechanism training to encourage noise robustness at privatisation time. These noised representations are then passed through a MLP, to give $\tilde{v}_i = g(z_i + s_i) = W^{(2)}\sigma(W^{(1)}(z_i+s_i))$, where $s_i\sim \text{Laplace}(0,b\mathbb{I})$.

We then maximise a softmax over cosine similiarities $\text{sim}(v,w) = v^T w / ||v||\hspace{0.4mm}||w||$ between positive pairs $(i,j)$:
\begin{align}
\ell_{ij} = -\log \frac{\exp \big( \text{sim}(\tilde{v}_i,\tilde{v}_j) \big)}{\sum_{k=1}^{2B}\text{sim}(\tilde{v}_i,\tilde{v}_k)} 
\end{align}
This encourages a large similarity between positive pairs and small similarity between negative pairs.

Since robustness to Laplace noise equates to large separation in L1 distance, we also experimented with L1-based similarity metrics. Empirically however, we found a significant decrease in representation quality (in the absence of noise) compared with cosine similarity. The best trade-off between representation quality and noise-robustness was achieved using cosine similarity with additive noise in representation space during training, as with the V-RLM. We refer to this method as the Contrastive RLM (C-RLM).


\subsection{Downstream Model Training on LDP Data}
\label{Section:downstream_training}

Having introduced a powerful approach for collecting high-dimensional data under LDP, we now introduce an approach for training downstream machine learning models using the collected LDP representations. This is a primary motive for data collectors, and furthermore, downstream model performance provides a powerful proxy for measuring the utility in our collected LDP data. We demonstrate this in the context of training classifiers, so we must also collect and privatise a discrete scalar label $y\in \{1,\ldots,K\}$ alongside our representation $\tilde{z}$. To privatise $y$, we flip it with probability
\begin{equation}
    p(\tilde{y} = i | y=j)
    = \frac{e^{\epsilon_y}\mathbb{I}(i=j)}{e^{\epsilon_y} + K - 1} 
    + \frac{\mathbb{I}(i\neq j)}{e^{\epsilon_y} + K - 1} 
\end{equation}
which induces $\epsilon_y$-LDP (see \citep{Wang2014} for proof). If $y \in \mathbb{R}$ (e.g. for regression) one could instead privatise this with, say, a Laplace Mechanism. By the composition theorem \citep{Dwork2014}, the tuple $(\tilde{z}, \tilde{y})$ satisfies $\epsilon$-LDP where $\epsilon = \epsilon_x + \epsilon_y$. Downstream models may be more robust to label noise than feature noise, or vice versa, so for fixed $\epsilon$ we set $\epsilon_x = \lambda\epsilon$ and $\epsilon_y = (1-\lambda)\epsilon$, with $\lambda$ chosen to optimise downstream model performance. 

To train a classifier with our collected LDP training data tuples $\mathcal{D}_2$, we assume the mechanism training data $\mathcal{D}_1$ follows a similar distribution to $\mathcal{D}_2$. We then place a prior $p(z|\mathcal{D}_1) = \frac{1}{M}\sum_{m=1}^M \delta(z - z'_m)$ over the (clean) representation space $z$ of our collected data $\mathcal{D}_2$, where $\mathcal{D}_1 := \{x'_m\}_{m=1}^M$ and $z'_m = f_\phi(x'_m)$. We then marginalise out both label and representation noise, and maximise
\begin{align}
    p_\psi(\tilde{y}, \tilde{z} | \mathcal{D}_1)
    &= \int \sum_{y} p(\tilde{y} | y) p_\psi(y | z)p(\tilde{z}|z)p(z|\mathcal{D}_1) \;dz \\
    &= \frac{1}{M}\sum_{m=1}^M \int \sum_{y} p(\tilde{y} | y) p_\psi(y | z)p(\tilde{z}|z) \delta(z - z'_m) \;dz \\
    &= \frac{1}{M}\sum_{m=1}^M \sum_{y} p(\tilde{y} | y) p_\psi(y| z'_m)p(\tilde{z}|z'_m) \label{eqn:clean_objective}
\end{align}
We can use $p_\psi(y|z)$ to classify \emph{clean} (i.e. non-privatised) data at inference time. To classify \textit{privatised} data, we only marginalise label noise and train $p_{\xi}(y|\tilde{z})$ by maximising:
\begin{align}
    p(\tilde{y}|\tilde{z}) = \sum_y p(\tilde{y}|y) p_{\xi}(y|\tilde{z}) \label{eqn:noisy_objective}
\end{align}
We find empirically that $p_{\xi}(y|\tilde{z})$ can also achieve a high accuracy classifying \emph{clean} representations. Though it is generally outperformed by $p_\psi(y|z)$, it can achieve better performance in scenarios where we don't have access to a good prior $p(z|\mathcal{D}_1)$. We hyperparameter tune our classifier using an LDP validation approach which we introduce in Appendix~\ref{appendix:private_grid_approach}.

The classification of both clean and private data are challenging problems with a multitude of applications. To our knowledge, no existing work has achieved compelling results on either problem in the high-dimensional setting. Figure \ref{fig:vlm_diagram} outlines a schematic of mechanism training, data privatisation, collection, and classifier training.

\section{Applications and Experiments}
\label{Section:experiments_clean}



The fundamental goal of an LDP mechanism is to maximise the retention of data utility while guaranteeing $\epsilon$-LDP. There are many ways to measure utility; we train downstream models on our LDP data, and use downstream model performance metrics as a proxy to measure the utility retained in our LDP training data. We conduct experiments on CIFAR-10 \citep{Krizhevsky09}, MNIST \citep{lecun1998}, and Lending Club\footnote{\url{https://www.kaggle.com/wordsforthewise/lending-club} (License CC0: Public Domain)} -- a tabular, binary classification task (see Appendix \ref{appendix:experimental_setup} for task details). In Sections \ref{Section:data_collection} and \ref{Section:novel_class_classification}, we train classifiers on LDP data to classify \emph{clean} (i.e.~non-privatised) datapoints at inference time. In Section \ref{Section:experiments_private}, we train a classifier on LDP data that classifies LDP datapoints at inference time, and refer to the performance metric of this separate use case as \emph{private} classification accuracy.

\textbf{Benchmarks:} We benchmark results against the Laplace mechanism, Duchi's mechanism \cite{Duchi2018} and the $\text{PrivUnit}_2$ mechanism \cite{Bhowmick2019} (see Appendix \ref{appendix:benchmark_setup} for implementation details). To our knowledge, the latter two represent current state-of-the-art. We discuss in Section~\ref{Section:Intro} why we do not benchmark against LDP mechanisms from \cite{Ding2017,Erlingsson2014,Fanti2016,Ren2018}. Classifiers trained with DP-SGD \cite{Abadi2016} do not constitute meaningful benchmarks since they provide no mechanism for LDP data collection, and require access to a non-LDP training set. While we assume access to non-LDP auxiliary data, this need not be labelled, or from the same distribution as our collected training data, as we demonstrate in Section~\ref{Section:novel_class_classification}.

\input{tables/data_collection}

\subsection{Data Collection} 
\label{Section:data_collection}

A fundamental objective of this paper is for organisations to utilise clean, auxiliary data to significantly improve the future, \emph{private} collection of user data. Vast numbers of organisations collecting data have access to existing datasets already e.g. public health bodies have access to medical images, tech companies have access to user data, and multinationals may have access to certain data from users in some regions but not others. In addition, there exist a broad array of public datasets for training machine learning models, and scraping data from the internet has become commonplace for training unsupervised models. Yet organisations still want to collect more data; this may be data from a group of patients in a particular study, e-commerce data from a broader group of users, or multinational data from a region where legislation is stricter, and so only private collection is acceptable.

In this section, we run experiments on CIFAR-10, MNIST, and Lending Club. As outlined in Appendix \ref{appendix:data_collection_split}, we split the data into an `auxiliary' dataset $\mathcal{D}_1$ used to train the mechanism and an external dataset $\mathcal{D}_2$ that we collect and privatise with said mechanism. This setup assumes $\mathcal{D}_1$ and $\mathcal{D}_2$ come from similar distributions, though in practice the distributions may differ. For example, $\mathcal{D}_2$ could be sales data collected in a different time period to $\mathcal{D}_1$, or user data collected in a different region. In Section \ref{Section:novel_class_classification}, we study such distributional shift, with $\mathcal{D}_2$ containing classes unseen in $\mathcal{D}_1$.  

Since a key motivation for organisations collecting data is to train machine learning models, we test the utility of the privatised $\mathcal{D}_2$ by using it to train classifiers. Results are shown in Table \ref{tab:data_collection}. For CIFAR-10, classifiers trained on $\epsilon$-LDP data privatised with the C-RLM outperform those trained on data privatised with benchmark mechanisms at every $\epsilon$ value tested. The C-RLM consistently achieves more than 50 percentage points higher than the benchmarks for $\epsilon\geq 6$, and achieves above random accuracy at $\epsilon \geq 2$. For MNIST and Lending Club we use the V-RLM, since data dimension is smaller, and the C-RLM's colour image augmentations are not applicable. Again, the classifier trained on $\epsilon$-LDP data privatised with the V-RLM significantly outperforms the benchmarks at every $\epsilon$ tested. These results indicate our mechanisms facilitate much greater retention of data utility than the benchmarks. ``No privacy'' indicates performance when no noise is added; for the benchmarks, this means training the classifier directly on non-private data $\mathcal{D}_2$, whilst for the V-RLM and C-RLM, it means training it on the non-private V-RLM and C-RLM representations of datapoints in $\mathcal{D}_2$.

In some scenarios, the auxiliary data $\mathcal{D}_1$ may contain sensitive information about individuals. In Appendix \ref{appendix:cdp_data_collection}, we train a V-RLM under CDP to protect the privacy of members of $\mathcal{D}_1$, demonstrating that the V-RLM still significantly outperforms benchmarks. In real-world scenarios, the size of $\mathcal{D}_1$ and $\mathcal{D}_2$ would be pre-determined by the amount of auxiliary data the organisation has access to, and the amount of data they are able to collect, respectively. We conduct an ablation study in Appendix \ref{appendix:datasplit_ablation} that shows our approach continues to outperform the benchmarks when the size of $\mathcal{D}_1$ is significantly reduced. In Appendix \ref{appendix:PCA_experiments}, we use PCA, rather than the (non-linear) V-RLM, to dimensionally reduce the data. Results suggest that not only are PCA representations less useful for downstream model training in the absence of LDP-inducing noise, they are also less robust to this noise than V-RLM representations. We attribute the robustness of V-RLM representations to the addition of Laplace noise during training.

\input{tables/novel_class}
\input{tables/data_join}

\subsection{Novel-Class Classification}
\label{Section:novel_class_classification}

As discussed in Section \ref{Section:data_collection}, the auxiliary data $\mathcal{D}_1$ and the data to be collected $\mathcal{D}_2$ may follow different data distributions. In the extreme case, the desired task on $\mathcal D_2$ may be to predict membership in a class that is not even present in dataset $\mathcal{D}_1$. For example, in a medical application there may be a large existing dataset of chest scans $\mathcal{D}_1$, but a public health body may want to collect data $\mathcal{D}_2$ from patients with a novel disease in order to train a novel-disease classifier to distribute to hospitals. Similarly, a software developer may have access to an existing dataset $\mathcal{D}_1$, but want to predict software usage data for $\mathcal{D}_2$, whose label is specific to the UI of a new release.

We run this experiment on MNIST and CIFAR-10, where the auxiliary $\mathcal{D}_1$ contains training images from classes 0 to 8, (with a small number of images held out for classifier training), and $\mathcal{D}_2$ contains all training images from class 9. See Appendix \ref{appendix:novel_class_split} for details. As in Sections \ref{Section:rep_learning} and \ref{Section:downstream_training}, we first train our mechanisms on $\mathcal{D}_1$, then privatise all images in $\mathcal{D}_2$ (we do not collect labels since all collected images have the same label). We then train a binary classifier on the dataset formed of the private 9's and the held out auxiliary images from classes 0-8 (which we privatise and label `not 9's'). Since $\mathcal{D}_1$ contains no datapoints from class 9, the prior from Equation \ref{eqn:clean_objective} is no longer as accurate. Consequently, as discussed in Section \ref{Section:downstream_training}, we found training our classifier with the objective from Equation \ref{eqn:noisy_objective} led to better performance than optimising Equation \ref{eqn:clean_objective}.

Results are shown in Table \ref{tab:hard_dist_shift}. Again, we outperform the benchmarks at all $\epsilon$ values tested, despite PrivUnit and Duchi's mechanism both performing significantly better on this simpler binary classification task than in the 10-class classification of Section \ref{Section:data_collection}. Notably, the C-RLM performs at 82.8\% accuracy at $\epsilon=1$, only 6.5\% less than at $\epsilon=10$.

\subsection{Data Joining}
\label{Section:data_joining}

\input{tables/data_collection_private}
An organisation training a classifier on some labelled dataset $\mathcal{D}_1$ could potentially improve performance by augmenting their dataset with other informative features, and so may want to join $\mathcal{D}_1$ with features from another dataset $\mathcal{D}_2$. We assume the owner of $\mathcal{D}_2$ may only be willing to share a privatised version of their dataset. For example, two organisations with mutual interests, such as the tax authorities and a private bank, or a fitness tracking company and a hospital, may want to join datasets to improve the performance of their algorithms. Similarly, it may be against regulations for multinational organisations to share and join non-privatised client data between departments in different regions, but permitted when the shared data satisfies LDP. 

We run this experiment on Lending Club. We split the datasets such that both $\mathcal{D}_1$ and $\mathcal{D}_2$ contain all rows, but $\mathcal{D}_1$ contains a subset of (clean) features, along with the clean label, and $\mathcal{D}_2$ contains the remaining features (to be privatised), as described in Appendix \ref{appendix:data_join_split}. 

We follow a privatisation procedure similar to that of Section \ref{Section:rep_learning}, with the distinction that the mechanism should be both trained on $\mathcal{D}_2$, and used to privatise $\mathcal{D}_2$. For the classification problem, instead of Equation~\ref{eqn:clean_objective} or \ref{eqn:noisy_objective}, we optimise $\log p_\psi(y_1|x_1, \tilde{x}_2)$ where $(x_1,y_1)\in\mathcal{D}_1$ and $\tilde{x}_2 \in \mathcal{D}_2^{\text{LDP}}$. We have access to all raw data needed for validation, eliminating the need to conduct a private grid as in previous experiments. Note that unlike the previous two experiments, we train the classifier on a combination of both clean and privatised features, and we classify this same combination of clean and privatised features at inference time.

Results are shown in Table \ref{tab:data_join}. The baseline of 56.1\% is the classification accuracy when using features from $\mathcal{D}_1$ only. Meanwhile, classifying on all (clean) features gives a 65.8\% accuracy. Neither benchmark achieves more than 1 percentage point accuracy increase over a classifier acting on the features from $\mathcal{D}_1$  only, whereas the V-RLM achieves a significant improvement for local $\epsilon \in [4,10]$.

We note that, unlike in previous experiments, we do not require auxiliary data to train the mechanism here. Given a dataset $\mathcal{D}$ that we want to privatise and share, we can both train a mechanism using $\mathcal{D}$, and use that mechanism to privatise $\mathcal{D}$. This can be viewed as a specific case of the more general problem of private data release. However, unlike CDP synthetic data generation approaches \citep{Acs2019, Takagi2021, Triastcyn2018, Xie2018}, each datapoint refers to a specific individual.

\section{Classifying Private Datapoints}
\label{Section:experiments_private}

In Sections \ref{Section:data_collection} and \ref{Section:novel_class_classification}, we investigated the use of LDP training data to train algorithms that classify clean datapoints. In some use cases however, we may want to train algorithms that act directly on LDP datapoints at inference time. Most notably, in the data collection framework, the organisation may want to do inference on individuals whose data they have privately collected.

However from the definition of LDP in Equation~\ref{eqn:LDP}, it is clear that a considerable amount of information is lost after privatisation, limiting classification accuracy. We would like to know the maximum achievable accuracy of a $K$-class classifier acting on representations privatised with an $\epsilon$-LDP local Laplace mechanism. Though challenging to calculate this maximum accuracy in general, we consider a simplified setting in Appendix~\ref{appendix:accuracy_bound_proof} (for representation dimension $d \geq K/2$), which forms an approximation to the maximum achievable accuracy given by:
\begin{align}
    \sum_{j=0, j \neq 1}^{K/2-1} \left( \binom{K/2-1}{j} \frac{(-1)^j}{1-j} \left[ \frac{e^{-j \epsilon /2}}{1+j}- \frac{e^{-\epsilon/2}}{2}  \right] \right) -\frac{\epsilon+1}{8}\left(K-2 \right) e^{-\epsilon/2}  \label{eqn:upper_bound}
\end{align}
In Table \ref{tab:data_collection_noisy}, we show the accuracy of classifiers when applied to privatised datapoints at inference time, and compare to this approximate maximum in Equation~\ref{eqn:upper_bound}. $\epsilon_\text{test}$ refers to the LDP guarantee of the images classified at test time, $\epsilon_\text{train}$ refers to the LDP guarantee of the (image, label) tuples used to train the classifier (where $\lambda = 0.7$). We see a drop in performance when classifying privatised datapoints, compared with results from Section \ref{Section:data_collection}. While we are not saturating the maximum from Equation~\ref{eqn:upper_bound}, we note that our method aims to build a downstream-task-agnostic, privatised representation of the data. Thus the representation must contain more information than just the class label. Meanwhile, Equation~\ref{eqn:upper_bound} is derived from the extreme setting in which the representation encodes only class information, and would be unable to solve any other downstream task. 

\section{Conclusion and Future Work}
\label{Section:conclusion}
In this paper we introduce a framework for collecting high-dimensional data under LDP. This is the first use of representation learning for LDP data collection, and we show that our approach overcomes the significant hurdles of privatisation in high-dimensions. Our approach can be easily adapted to any data type for which representation learning is possible. We demonstrate a range of applications, spanning important issues such as medical diagnosis, financial crime detection, and customer experience improvement, significantly outperforming existing baselines throughout.

\section*{Acknowledgments}
This work was developed and experiments were run on the Faculty Platform for machine learning. Alex Mansbridge and David Barber were supported by the Alan Turing Institute under the EPSRC grant EP/N510129/1.

\bibliography{references}
\bibliographystyle{icml2022}

\newpage
\appendix
\newpage
\section{Proof that the Local Laplace Mechanism Satisfies LDP} \label{appendix:laplace_mech_proof}
\textbf{Claim:} The local Laplace mechanism satisfies $\epsilon$-local differential privacy.
\\ \\
\textit{Proof:} We follow an approach similar to the proof in \citep{Dwork2014} that the central Laplace Mechanism satisfies CDP. Assume $x$ and $x'$ are two arbitrary datapoints. Denote $\mathcal{M}(x) = f(x) + (s_1, \ldots, s_k)$ where $s_i\sim \text{Laplace}(0, \Delta f / \epsilon)$. Then for some arbitrary $c$ we know that
\begin{align}
    \frac{p(\mathcal{M}(x) = c)}{p(\mathcal{M}(x') = c)} &=\prod_{i=1}^k\frac{p(\mathcal{M}_i(x) = c_i)}{p(\mathcal{M}_i(x') = c_i)}\\
    &= \prod_{i=1}^k\frac{e^{-\frac{\epsilon |f_i(x) - c_i|}{\Delta f}}}{e^{-\frac{\epsilon |f_i(x') - c_i|}{\Delta f}}} \\
    &= \prod_{i=1}^k e^{\frac{\epsilon (|f_i(x') - c_i| - |f_i(x) - c_i|)}{\Delta f}}\\
    &\leq \prod_{i=1}^k e^{\frac{\epsilon |f_i(x') - f_i(x)|}{\Delta f}}\\
    &= e^{\frac{\epsilon ||f(x') - f(x)||_1}{\Delta f}}\\
    &\leq e^\epsilon
\end{align}
where the first inequality comes from the triangle inequality, and the second comes from the definition of $\Delta f$. 

\section{Proof that Decoded Private Latents Satisfy LDP} \label{appendix:LDP_post_processing}
\textbf{Claim:} If a point in latent space satisfies $\epsilon$-LDP, then this point still satisfies $\epsilon$-LDP after being passed through a deterministic function, such as the function that parameterises the mean of the decoder network.
\\ \\
\textit{Proof:} We follow an approach similar to the proof that central differential privacy is immune to post-processing \citep{Dwork2014}. Let $\mathcal{A}: \mathcal{X} \rightarrow \mathcal{Z}$ be a randomised algorithm that satisfies $\epsilon$-LDP and $g: \mathcal{Z} \rightarrow \mathcal{Z}'$ be an arbitrary deterministic mapping. Let $S \subseteq \mathcal{Z}'$ and $T=\{z\in\mathcal{Z}:g(z)\in S\}$. Then
\begin{align}
    p\big(g(\mathcal{A}(x)) \in S\big) &= p\big(\mathcal{A}(x) \in T\big)\\
    &\leq e^\epsilon \, p\big(\mathcal{A}(x') \in T\big)\\
    &= e^\epsilon \, p\big(g(\mathcal{A}(x')) \in S\big)
\end{align}



\section{Hyperparameter Tuning Under LDP}
\label{appendix:private_grid_approach}
Typically for model validation and testing, one uses clean labels $y$ (and clean data $x$ when validating a classifier acting on clean data at inference time). The data collector does not have access to this, but we note they need only collect privatised model performance metrics on test/ validation sets, rather than accessing clean datapoints.

To do this, the trained classifier is sent to members of a validation/ test group, who would determine whether the classifier was correct $c \in \{0, 1\}$ on their data. Validation set members then return an $\epsilon$-LDP version $\tilde{c} \in \{0, 1\}$, flipped with probability $p = 1/(e^{\epsilon} + 1)$. The true validation set accuracy $A = \frac{1}{N_{\text{val}}}\sum_{n=1}^{N_{\text{val}}}c_n$  can be estimated from the privatised accuracy $\tilde{A} = \frac{1}{N_{\text{val}}}\sum_{n=1}^{N_{\text{val}}}\tilde{c}_n$ using $A = (\tilde{A} - p)/(1-2p)$ \citep{Warner1965}. We use this method when conducting a grid search over hyperparameters of our model, and to determine when to stop training.

\section{Experimental Setup} 
\label{appendix:experimental_setup}
For every experiment in the paper, we conduct three trials, and calculate the mean and standard deviation of accuracy for each set of trials. The error bars represent one standard deviation above and below the mean.

We use the CIFAR-10, MNIST and the Lending Club dataset. CIFAR-10 is a colour image dataset containing 60,000 images from 10 classes; MNIST is a dataset containing 70,000 grayscale images of handwritten digits from 10 classes, corresponding to digits 0-9. Lending Club is a tabular, financial dataset made up of around 540,000 entries with 23 continuous and categorical features (after pre-processing, before one-hot encoding); the task is binary classification, to determine whether a debt will be re-paid.

\subsection{Data Pre-Processing}
\label{appendix:data_preprocessing}
For MNIST and CIFAR-10, we convert the images to values between 0 and 1 by dividing each pixel value by 255 and treating them as continuous.

For Lending Club, a number of standard pre-processing steps are performed, including:
\begin{itemize}
    \item Dropping features that contain too many missing values, and those that would not normally be available to a loan company.
    \item Mean imputation to fill remaining missing values.
    \item Standard scaling of continuous features. Extreme outliers (those with features more than 10 standard deviations from the mean) are removed here.
    \item Balancing the target classes by dropping the excess class 0 entries.
    \item One-hot encoding categorical variables.
\end{itemize}
The target variable denotes whether the loan has been charged off or not, resulting in a binary classification task. The train, validation, test split is done chronologically according to the feature `issue date'. 

Note that in real world applications, the sizes of the RLM training/validation set, and the classifier training/validation sets would be pre-determined. For our experiments we use the data splits outlined in the following sections.

\subsubsection{Data Collection}
\label{appendix:data_collection_split}
\textbf{MNIST/ CIFAR-10:} The MNIST dataset contains 60,000 training points and 10,000 test points. CIFAR-10 contains 50,000 training points and 10,000 test points. For all results except those in Appendix \ref{appendix:datasplit_ablation}, we split both sets using 75\% for RLM training and the remainder for the classifier. The mechanism test set is used for validation since no test set is required here. The classifier training points are split randomly in a 9:1 ratio to form training and validation sets. We report classifier performance on the classifier test set.

\textbf{Lending Club:} This dataset is split into train, validation and test sets according to the issue date of the loans. The oldest 85\% of data forms the training data, with the remaining forming the validation and test data. As with MNIST and CIFAR-10, we use 75\% for the split between RLM training data and classifier training data.

We emphasise that we split these datasets to simulate real world scenarios. In reality, an organisation using this approach would construct a RLM training set from clean data they have access to, either internally or publicly. That organisation can then use this trained RLM to collect further data under LDP guarantees. We use this collected private data to train a classifier, but it could be used for many tasks. In Appendix \ref{appendix:datasplit_ablation} we explore the extent to which the size of the RLM training set $\mathcal{D}_1$ affects classifier performance.

\subsubsection{Novel-Class Classification}
\label{appendix:novel_class_split}

\textbf{MNIST/ CIFAR-10:} We use a similar approach to the above, but split the data between the mechanism and the classifier such that the mechanism train/validation sets contain $\frac{8}{9}$ths of (unlabelled) training images from classes 0 to 8. The remaining $\frac{1}{9}$th of 0 to 8 images, and all 9s, are used for classifier train, test and validation sets. Our mechanism datasets then contain equal class balance for the classes 0 to 8, and the classifier datasets contain equal class balance for 9s and `not 9s'.

\subsubsection{Data Join}
\label{appendix:data_join_split}
\textbf{Lending Club:} For this experiment, the datasets are split column-wise, between the dataset’s 23 features, such that 8 features remain non-privatised (month of earliest credit line, number of open credit lines, initial listing status of loan, application type, address (US state), home ownership status, employment length, public record bankruptcies) and the remaining 15 features are privatised. This feature split was chosen such that the 8 non-private features contain some information to solve the classification task, while the remaining 15 features contain information which, at least before privatisation, further improves classifier performance.

\subsection{Benchmarks}
\label{appendix:benchmark_setup}
We studied the local Laplace mechanism, Duchi's mechanism \cite{Duchi2018}, the Hybrid mechanism \cite{Wang2019}, and the $\text{PrivUnit}_2$ mechanism \cite{Bhowmick2019}.

For the Laplace mechanism, we privatise each feature independently, choosing the noise level for each of the $d$ features such that $\epsilon_i = \epsilon/d$, We do this since we have no prior knowledge about which features are most important. We then assume $\Delta f_i$ is equal to the difference between the maximum and minimum value of the feature $i$ within the training and validation sets used to train the mechanism in the main experiments, after pre-processing. One then has to clip any values that lie outside this interval in the collected dataset at privatisation time. For tabular experiments, we privatise categorical features using a flip mechanism. 

We omit the hybrid mechanism (Algorithm 4 of \citet{Wang2019}) since it entails collecting only $k\leq4$ of the $d$ features for our experiments, but implement Duchi's mechanism as in Algorithm 3 of \citet{Wang2019}. Again, we privatise categorical features using a flip mechanism.

For $\text{PrivUnit}_2$, treat each image as a $d$-dimensional vector and privatise both its direction and magnitude, as outlined in the $\text{PrivUnit}_2$ and ScalarDP algorithms in \cite{Bhowmick2019}. We omit Lending Club experiments, since it is not designed for mixed-type data.

For the CIFAR-10 experiments, we experimented with using both a pre-trained and a randomly initialised ResNet-18 for classification, as is used in the C-RLM encoder, but found a feedforward architecture with hidden layers of dimension \{400, 150, 50\} achieved the best accuracy at all local-$\epsilon$ values (except local-$\epsilon = \infty$, where instead a pre-trained ResNet-18 architecture was used).

For MNIST and Lending club experiments, we used only feedforward architectures owing to the smaller data dimension.

\subsection{Hyperparameter Choices}
\label{appendix:hyperparameter_choices}
In order to find the optimal experimental setup, we conducted a grid search over a number of the hyperparameters in our model.

For V-RLM training, we use a learning rate of $10^{-4}$ and batch size of 128 for Lending Club experiments, and we use a learning rate of $5\times 10^{-4}$ and batch size of 64 for MNIST. For the C-RLM training we use a learning rate of $3\times10^{-4}$ and a batch size of 128 (the largest our GPU would allow). We searched over the following model hyperparameters:
\begin{itemize}
\vspace{-1mm}
    \item The proportion $\lambda$ of our privacy budget assigned to the representation vs. the label i.e. $\lambda=\epsilon_x / (\epsilon_x + \epsilon_y)$.
    \item The $\ell_1$ clipping distance $l$ of our inference network mean i.e. $l = \Delta f_\phi / 2$.
    \item The Laplace distribution scale $b$ of our approximate posterior distribution during pre-training of the RLM. Note that we report this in terms of the $\epsilon_\text{pre-training}$-LDP value induced by this Laplace noise i.e. $\epsilon_\text{pre-training} = 2l/b$. This is fixed throughout training, unless `learnt' (i.e. the parameter $b$ is a learnt scalar in the V-RLM)  or `no privacy' (i.e. $b=0$) is specified in Table \ref{Table:hyperparameters_data_collection}. 
    \item The representation dimension $d$. We fixed $d=32$ for CIFAR-10 and $d=8$ for MNIST. For Lending Club, we fixed $d=8$ for the data collection experiments but searched over $d\in\{5, 8\}$ for the data join experiments due to the smaller number of features.
\end{itemize}

For CDP training experiments in Appendix \ref{appendix:cdp_data_collection}, we also did a grid search over values for the noise multiplier, batch size, and DP learning rate for central $\epsilon\in \{1,5\}$. The DP-Adam \cite{Gylberth2017} hyperparameter max gradient norm was fixed to 1 throughout. The number of training epochs needed to reach the target central $\epsilon$ value follows from the choice of hyperparameters, combined with the V-RLM training set size (45,000 for MNIST, and 341,000 for Lending Club). Note that we fixed $\delta = 10^{-5}$ for all experiments.

The results from these grid searches are given in Tables \ref{Table:hyperparameters_data_join}, \ref{Table:dp_hyperparameters}, and \ref{Table:hyperparameters_data_collection}. 

\input{tables/hparams_datajoin}

\input{tables/hparams_dpadam}

\input{tables/hparams_mechanisms}

\subsection{Mechanism Architectures and Transformations}
\label{apendix:transformations}

For MNIST, we use a V-RLM encoder network with 3 hidden layers of size \{400, 150, 50\}, and a decoder network with 3 hidden layers of size \{50, 150, 400\}. For Lending Club, we use a V-RLM encoder and decoder network with 2 hidden layers of size \{500, 500\}. For the classifier, we used a a network with 1 hidden layer of size 50.

For CIFAR-10, we use a pre-trained ResNet-18 model for our encoder, followed by a final hidden layer of size 32. To classify the C-RLM representations, we used a simple logistic regression classifier. 

When training the C-RLM, the $B$ images in each minibatch are augmented twice via randomly sampled transformations, to form $2B$ positive pairs and $2(B-1)$ negative pairs. Following \citet{Chen2020}, we use the following transformations:
\begin{itemize}
    \item Random cropping and resizing to 224x224,
    \item Random flipping with probability 0.5,
    \item Colour jitter applied with probability 0.8,
    \item Conversion to grayscale with probability 0.2.
\end{itemize}

\textbf{Computational Cost:} All experiments in this paper were trained on a single NVIDIA RTX 2080-Ti GPU. For V-RLM experiments, training took under 1 hour, except when a two stage process was used to train the parameters of the encoder under CDP, as in Appendix~\ref{appendix:cdp_data_collection}. In this instance, training time varied significantly with DP-Adam hyperparameter choices, but took under 30 minutes for all MNIST experiments and under 3 hours for Lending Club. For C-RLM experiments, performance improved significantly with further training epochs, but no C-RLM was trained for more than 12 hours on a single GPU. To train the classifier on private data took under 10 minutes.

\section{Data Collection with Mechanisms Trained Under CDP} \label{appendix:cdp_data_collection}
\input{tables/data_collection_cdp}

We first formalise the definition of central differential privacy:

\textbf{Definition ($(\epsilon, \delta)$-central differential privacy):} Let $\mathcal{A}: \mathcal{D} \rightarrow \mathcal{Z}$ be a randomised algorithm, that takes as input datasets from the dataset domain $\mathcal{D}$. We say $\mathcal{A}$ is $(\epsilon, \delta)$-central differentially private if for $\epsilon, \delta \geq 0$, for all subsets $S \subseteq \mathcal{Z}$, and for all neighbouring datasets $D, D' \in \mathcal{D}$, we have
\begin{equation}
    p(\mathcal{A}(D) \in S) \;\leq\; e^\epsilon \, p(\mathcal{A}(D') \in S) + \delta
\end{equation}
where for $D$ and $D'$ to be neighbouring means that they are identical in all but one datapoint. 

Intuitively, this states that one cannot tell (with a level of certainty determined by $(\epsilon, \delta)$) whether an individual is present in a database or not.

In the scenario that the auxiliary dataset $\mathcal{D}_1$ contains sensitive information, the parameters of the encoder may need to satisfy CDP with respect to $\mathcal{D}_1$. The primary use-case for this would be when the encoder is shared with an untrusted third-party (i.e. the person from whom you wish to collect data). For the C-RLM model, we have only an encoder and this can be trained relatively easily using a private optimisation algorithm such as DP-Adam \cite{Gylberth2017}, but for the V-RLM, we found the following two-stage training approach to be effective:
\begin{itemize}
\vspace{-2.5mm}
    \item \textbf{Stage 1:} Train a V-RLM with encoding distribution $q_\phi(z|x)$ and decoding distribution $p_\theta(x|z)$ using a non-DP optimiser e.g. Adam \citep{Kingma2015}.
    \vspace{-1mm}
    \item \textbf{Stage 2:} Fix $\theta$ and re-initialise the encoder with a new distribution $q_{\phi_\text{private}}(z|x)$. Optimise $\phi_\text{private}$ using DP-Adam \citep{Gylberth2017}.
\end{itemize}
\vspace{-2.5mm}

We study the effects of training the V-RLM under CDP, by recreating the experiments from Section \ref{Section:data_collection} using a $(\epsilon_\text{CDP},\delta_\text{CDP})$-CDP encoder trained with the two-stage approach above. Results are shown in Table \ref{tab:cdp_data_collection}. For all stated $(\epsilon_\text{CDP},\delta_\text{CDP})$-CDP results we use $\delta_\text{CDP} = 10^{-5}$. 

For MNIST, we see that classifier accuracy deteriorates slightly when data is collected using mechanisms with lower $\epsilon_\text{CDP}$ values. This suggests that less information is contained in representations privatised with the CDP mechanisms, as expected. However, we still significantly outperform all benchmarks at all $\epsilon_\text{LDP}$ values.

For Lending Club, we see virtually no deterioration in classification accuracy when collecting data using a CDP encoder, which we attribute to the larger training set $\mathcal{D}_1$.


\section{Effects of Reducing Auxiliary Dataset Size} \label{appendix:datasplit_ablation}
\input{tables/ablation_datasplit}

We conduct an ablation study in order to determine the effect of the V-RLM training set size on the utility of data privatised with the mechanism. Specifically, we study the MNIST data collection experiment. For the MNIST experiments in Sections \ref{Section:experiments_clean} and \ref{Section:experiments_private}, we use 75\% of the data for training the V-RLM, and privatise the remainder to form the classifier training set. In this study, we train the V-RLM on different training set sizes. Specifically, we train mechanisms of proportions $\eta = \{75\%, 50\%, 25\%, 10\%\}$ of the MNIST training set, corresponding to 45,000, 30,000, 15,000, and 6,000 unlabelled training images respectively. 

Results, shown in Table \ref{tab:ablation_datasplit} show the accuracy of classifiers trained on data collected by each mechanism at a range of $\epsilon$-LDP guarantees. Note that these experiments use hyperparameters optimised for experiments in Section \ref{Section:experiments_clean} where the V-RLM used $\eta=75\%$ of the training data, and so performance on smaller V-RLM training sets could potentially be improved with hyperparameter tuning. 

We also experimented with different data splits for the benchmarks to determine whether this could improve the classifier performance. Since we do not require a pre-training set $\mathcal{D}_1$ for the benchmarks, the most favourable setup is to assign 100\% of the data (60,000 labelled images) to the classifier training set $\mathcal{D}_2$. All 3 benchmark results shown in Table \ref{tab:ablation_datasplit} use this favourable data split. Regardless, we see that none of the benchmarks can compete with any of the V-RLM models at any $\epsilon$-LDP guarantee.

\section{Comparision with Simpler Dimensionality-Reduction Approaches} \label{appendix:PCA_experiments}

\input{tables/data_collection_pca}

In this experiment, we aim to determine whether the utility level of data privatised with the V-RLM can be achieved with simpler dimensionality reduction techniques. We compare the V-RLM to a PCA-based LDP mechanism. The experimental set-up is the same as in the data collection experiment: we use the dataset $\mathcal{D}_1$ to learn the principal components, and use these to reduce our dataset $\mathcal{D}_2$ to dimension 8, as in the V-RLM (with data splits as described in Appendix \ref{appendix:data_collection_split}). We then add Laplace noise to privatise $\mathcal{D}_2$, and this privatised dataset is used to train a classifier. 

There are two differences between the PCA mechanism and the V-RLM: firstly, the mapping $f(\cdot)$ in Equation \ref{eqn:local_laplace_mechanism} is linear for PCA, while it is a non-linear neural network in the V-RLM; secondly, the V-RLM adds Laplace noise during training, and so the representations should be more robust to privatisation noise than the data reduced with PCA.

The results are shown in Table \ref{tab:data_collection_pca}. Notably, we see that when no noise is added to the representations (i.e. $\epsilon=\infty$), the PCA-based mechanism is outperformed by the V-RLM, suggesting the clean PCA representations contain less information than the V-RLM representations for downstream model training. Furthermore, performance drops significantly when any noise is added to the PCA representations (i.e. $\epsilon\leq10$), whilst for the V-RLM, the addition of noise has far less impact on classifier accuracy. This suggests that the PCA representations lack robustness to noise. Indeed at $\epsilon=10$, PCA is outperformed by the PrivUnit benchmark -- a technique that does not use dimensionality reduction.


\section{Proof of Upper Bound on Private Classification Accuracy} \label{appendix:accuracy_bound_proof}

\subsection{General Upper Bound}
\label{vlm_sec:general_model}
Recall we have some function $f(\cdot): \mathcal{X} \rightarrow \mathcal{Z}$ that maps data $x$ to representations $z = f(x)$ inside a $d$-dimensional taxicab sphere $\mathcal{Z}$ of diameter $\Delta f$. We pass $x$ through a randomised algorithm $\mathcal{M} = f(\cdot) + (s_1,\hdots, s_d)$ to induce LDP, where $s_i \sim \text{Laplace}(0, b)$. We denote the LDP point $\tilde{z}$. 

We make the simplifying assumption that we have equal class balance, which is the case for all experiments in this work. 

The classifier $\mathcal{C}$ will partition $\mathbb{R}^d$ into $\{S^{(1)},\hdots, S^{(K)}\}$ such that $\tilde{z}\in S^{(k)}$ will be classified into class $k$. The accuracy of $\mathcal{C}$ is then given by
\begin{align}
    A &= \mathbb{E}_{(x,y)\sim p(x,y), \tilde{z}\sim \mathcal{M}(\tilde{z}|x)} \bigg[\mathbb{I}\left(\tilde{z}\in S^{(y)}\right)\bigg]\\
    &= \mathbb{E}_{(x,y)\sim p(x,y)}\left[\int_{\tilde{z}\in \mathbb{R}^d} \mathbb{I}\left(\tilde{z}\in S^{(y)}\right)\frac{1}{(2b)^d} e^{-\frac{||f(x) - \tilde{z}||_1}{b}}d\tilde{z}\right]\\
    &= \mathbb{E}_{(x,y)\sim p(x,y)}\left[ \int_{\tilde{z}\in S^{(y)}} \frac{1}{(2b)^d} e^{-\frac{||f(x) - \tilde{z}||_1}{b}}d\tilde{z}\right]  \label{clm_eq:noisy_acc_1}
\end{align}

We observe that it is always possible to achieve a greater (or equal) classification accuracy if $f(\cdot)$ maps all data from a given class $k$ to a single point on the taxicab sphere, rather than some region $\Gamma^k\subseteq\mathcal{Z}$ containing more than one point.

To justify this, suppose all points from class $k$ are mapped, via $f(\cdot)$, to some region $\Gamma^k$. The classifier defines a decision region $S^{(k)}$ such that any point inside $S^{(k)}$ gets classified as class $k$. There exists (at least one) point $c^{(k)} \in \Gamma^{(k)}$ such that $\forall g^{(k)} \in \Gamma^{(k)}$:
\begin{align}
    \int_{\tilde{z}\in S^{(k)}} \frac{1}{(2b)^d} e^{-\frac{||c^{(k)} - \tilde{z}||_1}{b}}d\tilde{z} \geq \int_{\tilde{z}\in S^{(k)}} \frac{1}{(2b)^d} e^{-\frac{||g^k - \tilde{z}||_1}{b}}d\tilde{z}
\end{align}
This says that $c^{(k)} \in \Gamma^{(k)}$ is the point such that the $\text{Laplace}(c^{(k)}, b)$ distribution contains more probability mass inside our decision region $S^{(k)}$ than any other distribution of the form $\text{Laplace}(g^{(k)}, b)$ with $g^{(k)} \in \Gamma^{(k)}$. In other words, $c^{(k)}$ is the representation inside $\Gamma^{(k)}$ most likely to still be classified as class $k$ after privatisation. So if we modify $f(\cdot)$ such that all points from class $k$ are mapped to a single representation $c^{(k)}\in \Gamma^{(k)}$, we will achieve higher (or equal) accuracy than with the original $f(\cdot)$. 

In light of this, we assume $f(\cdot)$ maps all data from class $k$ to a single representation $c^{(k)}$ and write Equation~\ref{clm_eq:noisy_acc_1} as
\begin{align}
    A &= \mathbb{E}_{y\sim p(y)}\left[ \int_x \int_{\tilde{z}\in S^{(y)}} \frac{1}{(2b)^d} e^{-\frac{||f(x) - \tilde{z}||_1}{b}}d\tilde{z}\;p(x|y)dx\right]\\
    &= \mathbb{E}_{y\sim p(y)}\left[  \int_{\tilde{z}\in S^{(y)}} \frac{1}{(2b)^d} e^{-\frac{||c^{(y)} - \tilde{z}||_1}{b}}d\tilde{z} \right]\\
    &= \frac{1}{K}\sum_{y=1}^K \int_{\tilde{z}\in S^{(y)}} \frac{1}{(2b)^d} e^{-\frac{||c^{(y)} - \tilde{z}||_1}{b}}d\tilde{z} \label{clm_eq:noisy_acc_2}
\end{align}

where the last inequality follows from the assumption of equal class balance, and the integral inside the sum
\begin{equation}
    A^{(y)} := \int_{\tilde{z}\in S^{(y)}} \frac{1}{(2b)^d} e^{-\frac{||c^{(y)} - \tilde{z}||_1}{b}}d\tilde{z}
\end{equation}
represents the probability of correctly classifying a noised representation from class $y$, and is equal to the total probability mass of $\text{Laplace}(c^{(y)}, b\mathbb{I})$ inside $S^{(y)}$.

We know that the $A$ will be maximised when the decision boundaries are such that a noised representation $\tilde{z}$ is classified as coming from class $y$ if $c^{(y)}$ is the closest of the $K$ representations $\{c^{(1)},\hdots,c^{(K)}\}$ in L1-distance. This is because the probability mass at $\tilde{z}$ will be highest under a distribution with mean closest to $\tilde{z}$, and so $\tilde{z}$ will contribute more mass to $A^{(y)}$ than it would any other $A^{(\hat{y})}$ for $\hat{y}\neq y$. Thus $A=\frac{1}{K}\sum_{y=1}^K A^{(y)}$ will be higher. 

In L1 geometry there exist subsets of $\mathbb{R}^d$ where all points in the subset lie equidistant from multiple class representations, as shown by the red shaded regions in Figure~\ref{vlm_fig:equidistant_left}. In this scenario, the choice of decision boundary through such subsets will not affect accuracy -- assigning a region to class $y$ rather than some equidistant class $\tilde{y}\neq y$ will increase $A^{(y)}$ and reduce $A^{(\hat{y})}$ by the same amount, leaving $A$ unchanged. We can therefore arbitrarily assign noised representations on such hyperplanes to any of the closest (in L1-norm) classes - a simple approach is to assign them to the class representation closest in L2-norm, as shown in Figure~\ref{vlm_fig:equidistant_right}. 

\input{tables/fig_proof_equidistant}

In order to find the maximum achievable accuracy, we must find the representations $\{c^{(1)}, \hdots,c^{(K)}\}$ within $\mathcal{Z}$ that maximise Equation~\ref{clm_eq:noisy_acc_2}, where our decision regions $S^{(y)}$ are defined as described above. 

We note that $S^{(y)}$ correspond to Voronoi cells with generators $c^{(y)}$. Thus our problem is equivalent to finding the optimal positioning of generators such that the probability mass inside each cell (corresponding to the distribution centred at that cell's generator) is maximised. Finding the optimal generator locations has been widely studied in fields such as computational geometry \citep{Bhattacharya2010} and operations research \citep{Riol2011}. Studies of probability density in Voronoi cells have appeared in fields such as Astrophysics \citep{Jamieson2021}. 

Our task of finding the locations of the $K$ optimal generators inside the $d$-dimensional taxicab sphere is challenging, and we leave a detailed analysis to future work. Here, we instead study a simple setting which we hypothesise to a good approximation to the optimal solution, for the setting defined in our experiments.

\subsection{Simplified Setting}
\label{vlm_sec:simplified_model}

It seems reasonable to assume that when the taxicab sphere $\mathcal{Z}$ has at least as many vertices as there are classes (i.e. $d\geq K/2$, as in all experiments in this work), the representations $\{c^{(1)},\hdots,c^{(K)}\}$ that lead to the highest accuracy will lie on the vertices of $\mathcal{Z}$. Furthermore, when $d>K/2$, we hypothesise it is favourable to place representations opposite one another (i.e. $c^{(i)} \cdot c^{(j)} = -\Delta f^2 / 4$)) rather than on different axes (i.e. $c^{(i)} \cdot c^{(j)} = 0$), where possible.

Supporting this assumption, when $d=K=2$ it is straightforward to show that the accuracy is highest when $c^{(1)}$ and $c^{(2)}$ lie on opposite vertices rather than elsewhere on the boundary of $\mathcal{Z}$. Furthermore, numerical simulations of this problem similarly conclude that for $d\in\{2,3\}$ and $K\in\{2,\dots,6\}$, the optimal setting is to place the representations on vertices of the sphere (and indeed opposite vertices when $d>K/2$). We found higher dimensional problems to be too computationally expensive to simulate numerically. 

In light of this, we construct a setting that places representations on (opposite) vertices of $\mathcal{Z}$, which we expect to be a good approximation of the true maximum achievable accuracy. Let $e_i$ be the $i^\text{th}$ standard basis vector. Then we define our representations $\{c^{(1)}, \hdots,c^{(K)}\}$ as
\begin{equation}
    c^{(k)} = \begin{cases}
\frac{1}{2} \Delta f \cdot e_{\frac{k}{2}}, & \text{for $k$ odd}\\
-\frac{1}{2} \Delta f \cdot e_{\frac{k+1}{2}}, & \text{for $k$ even}
\end{cases}
\end{equation}
This places representations on opposite vertices of the taxicab sphere, for each axis in turn, until all representations have been assigned. Corresponding decision regions as defined as in Section~\ref{vlm_sec:general_model}. Figure~\ref{vlm_fig:decision_boundary} shows these representations and corresponding decision boundaries for $d=2$ and $K=4$ and $\Delta f = 4$. 

\input{tables/fig_proof_boundary1}

Given these representations, and the inferred decision regions, we can calculate the accuracy of the optimal classifier given by Equation~\ref{clm_eq:noisy_acc_2}. We first consider the case of $K=2d$, meaning we have the same number of vertices as classes. To ease notation we assume WLOG that $\Delta f = 2$, and so $c^{(1)} = (1,0,\hdots,0)$. We also denote $\tilde{z} = (\tilde{z}_1,\hdots,\tilde{z}_d)$. 

First, we note that by symmetry, we see that $A_1 = A_2 = \hdots = A_K$, and so $A = A_1$. The decision boundary $S^{(1)}$ is defined as 
\begin{align}
S^{(1)} = \{(\tilde{z}_1, \hdots \tilde{z}_d): \;\tilde{z}_1>0, \;\;\text{and}\;\; \tilde{z}_i<|\tilde{z}_1|,\; \forall i\in\{2, \hdots, K/2\}\}    
\end{align}

We can then calculate the accuracy $A$ as follows:
\begin{align}
A &= \int_{\tilde{z}\in S^{(1)}} \frac{1}{(2b)^d} e^{-\frac{||c^{(1)} - \tilde{z}||_1}{b}}d\tilde{z}\\
&= \int_{\begin{subarray}{l}\tilde{z}_1>0, \\ \tilde{z}_i<|\tilde{z}_1|, \forall i\neq1 \end{subarray}} \frac{1}{(2b)^d}e^{-\frac{||c^{(1)} - \tilde{z}||_1}{b}} d\tilde{z}_{1:d}\\
&=\int_0^\infty \frac{1}{2b}e^{-\frac{|1 - \tilde{z}_1|}{b}} \bigg(\prod_{i=2}^d \int_{-\tilde{z}_1}^{\tilde{z}_1}  \frac{1}{2b}e^{-\frac{|\tilde{z}_i|}{b}}d\tilde{z}_i \bigg)d\tilde{z}_1\\
&=\int_0^\infty \frac{1}{2b}e^{-\frac{|1 - \tilde{z}_1|}{b}} \bigg(1 - e^{-\tilde{z}_1/b}\bigg)^{d-1} d\tilde{z}_1\\
&=\int_0^\infty \frac{1}{2b}e^{-\frac{|1 - \tilde{z}_1|}{b}} \sum_{j=0}^{d-1} \binom{d-1}{j}(-1)^j e^{-\frac{j\tilde{z}_1}{b}} d\tilde{z}_1\\
&=\sum_{j=0}^{d-1} \binom{d-1}{j} (-1)^{j} \int_0^\infty \frac{1}{2b}e^{-\frac{|1 - \tilde{z}_1|}{b}}  e^{-\frac{j\tilde{z}_1}{b}}d\tilde{z}_1\\
&=\sum_{j=0}^{d-1} \binom{d-1}{j} (-1)^{j} \Bigg[\int_0^1 \frac{1}{2b} e^{-\frac{\tilde{z}_1(j-1) + 1}{b}} d\tilde{z}_1 + \int_1^\infty \frac{1}{2b} e^{-\frac{\tilde{z}_1(j+1) - 1}{b}} d\tilde{z}_1\Bigg]\\
&=\frac{1-d}{2b}\bigg( 1+\frac{b}{2}\bigg)e^{-1/b} + \sum_{j=0, j \neq 1}^{d-1} \binom{d-1}{j} (-1)^j \Bigg[ 
\frac{e^{-j/b}}{1-j^2} -\frac{e^{-1/b} }{2(1-j)} \Bigg]\\
&=(1-d)\frac{\epsilon+1}{4}e^{-\epsilon/2}  + \sum_{j=0, j \neq 1}^{d-1} \binom{d-1}{j} (-1)^j \Bigg[ 
\frac{e^{-j\epsilon/2}}{1-j^2} -\frac{e^{-\epsilon/2} }{2(1-j)} \Bigg]\\
&=\sum_{j=0, j \neq 1}^{K/2-1} \left( \binom{K/2-1}{j} \frac{(-1)^j}{1-j} \left[ \frac{e^{-j \epsilon /2}}{1+j}- \frac{e^{-\epsilon/2}}{2}  \right] \right) - \frac{\epsilon+1}{8}\left(K-2 \right) e^{-\epsilon/2}  \label{vlm_eqn:upper_bound}
\end{align}
where in the penultimate step we used the fact that for $\epsilon$-LDP we have $b=2/\epsilon$, and in the final equality we substitute $d=K/2$.

When $K \leq 2d$ and $K$ is even, we note that the accuracy is unchanged. The decision boundary $S^{(1)}$ is now defined as 
\begin{align}
S^{(1)} = \{(\tilde{z}_1, \hdots \tilde{z}_d): \;&\tilde{z}_1>0, \nonumber \\
&\tilde{z}_i<|\tilde{z}_1|,\; \forall i\in\{2, \hdots, K/2\},\nonumber\\
&\tilde{z}_j \in (-\infty, \infty), \; \forall j\in\{K/2 + 1, \hdots, d\}\}    
\end{align}
where the unbounded dimensions integrate to 1, leaving accuracy unchanged. Equation~\ref{vlm_eqn:upper_bound} therefore defines the maximum achievable accuracy in this simplified setting, and a proxy for the maximum achievable accuracy of the private classifiers studied in this work. We omit the case of $K \leq 2d$ and $K$ odd, since $K$ is even in all experiments.

\end{document}

%% file: tikzstuff.tex
\usepackage{tikz}
\usetikzlibrary{arrows,shapes,backgrounds,through,shadows}

\tikzstyle{cont}=[circle, draw,thick,minimum size=7.5mm,line width=1pt,>=stealth]  
\tikzstyle{contbig}=[circle, draw,thick,minimum size=8.5mm,line width=1pt,>=stealth]  
\tikzstyle{disc}=[diamond, draw,thick,minimum size=7.5mm,line width=1pt,>=stealth]  
\tikzstyle{obs}=[fill=blue!10,thick]  

\tikzstyle{obs}=[fill=blue!10,thick]  

\tikzstyle{contobs}+=[cont]
\tikzstyle{contobs}+=[obs]
\tikzstyle{discobs}+=[disc]
\tikzstyle{discobs}+=[obs]
\tikzstyle{contobsbig}+=[contbig]
\tikzstyle{contobsbig}+=[obs]
\tikzstyle{dgraph}=[->, line width=1.5pt]

%% file: math_commands.tex
\usepackage{amsmath,amsfonts,bm}

\def\eqref#1{equation~\ref{#1}}

\def\1{\bm{1}}

\def\eps{{\epsilon}}

\DeclareMathAlphabet{\mathsfit}{\encodingdefault}{\sfdefault}{m}{sl}
\SetMathAlphabet{\mathsfit}{bold}{\encodingdefault}{\sfdefault}{bx}{n}

\newcommand{\KL}{D_{\mathrm{KL}}}



%% file: tables/data_collection.tex
\begin{table*}[t]
\caption{Accuracy of classifiers trained on data collected using different LDP mechanisms. Each column shows the $\epsilon$-LDP guarantee for the collected training set. Error bars represent $\pm 1$ standard deviation from the mean over 3 trials.}
\label{tab:data_collection}
\vspace{-4mm}
\begin{center}
\begin{small}
\begin{sc}
\begin{tabular}{p{1mm}p{22mm}|p{11mm}p{11mm}p{11mm}p{11mm}p{11mm}p{11mm}|p{13mm}}
\toprule
& \makecell[l]{Mechanism}
& \makecell[c]{$\eps=10$}
& \makecell[c]{$\eps=8$}
& \makecell[c]{$\eps=6$}
& \makecell[c]{$\eps=4$}
& \makecell[c]{$\eps=2$}
& \makecell[c]{$\eps=1$}
& \makecell[c]{No LDP}
\\
\midrule
&C-RLM (Ours)
&\textbf{75.9±3.8}
&\textbf{75.3±1.5}
&\textbf{73.9±3.4}
&\textbf{43.4±11.8}
&\textbf{17.6±3.9}
&\textbf{15.0±4.0}
&86.3±0.0
\\ 
\multirow{3}{*}{\rotatebox[origin=c]{90}{\parbox[c]{1cm}{\centering \;CIFAR10}}} 
&PrivUnit
&18.9±1.1
&13.2±2.9
&11.0±2.0
&9.6±0.4
&10.1±0.1
&9.5±0.5
&76.1±0.7
\\ 
&Duchi
&12.8±1.1
&11.1±0.7
&10.6±0.7
&10.9±0.9
&10.4±0.8
&9.9±0.1
&76.1±0.7
\\ 
&Laplace
&10.2±0.4
&9.9±0.4
&9.9±0.1
&9.6±0.5
&10.0±0.1
&9.8±0.1
&76.1±0.7
\\ \hline
&V-RLM (Ours)
&\textbf{86.1±1.0}
&\textbf{82.1±1.6}
&\textbf{72.8±3.2}
&\textbf{61.3±2.8}
&\textbf{35.3±9.7}
&\textbf{16.9±1.3}
&94.9±0.2
\\ 
\multirow{3}{*}{\rotatebox[origin=c]{90}{\parbox[c]{1cm}{\centering \;\;MNIST}}}
&PrivUnit
&38.2±4.6
&15.1±4.2
&12.4±1.6
&9.2±2.6
&9.5±2.9
&11.3±4.5
&96.0±0.4
\\ 
&Duchi
&13.9±4.5
&14.3±5.0
&13.3±3.8
&14.1±4.9
&14.0±5.6
&10.0±1.1
&96.0±0.4
\\ 
&Laplace
&\;\;9.2±1.4
&\;\;9.8±2.8
&10.6±0.3
&10.0±0.6
&\;\;9.0±1.1
&10.0±0.8
&96.0±0.4
\\ \hline
&V-RLM (Ours)
&\textbf{63.7±0.6}
&\textbf{63.2±0.3}
&\textbf{62.6±0.5}
&\textbf{61.1±1.2}
&\textbf{53.9±4.3}
&\textbf{55.1±4.5}
&65.0±0.3
\\ 
\multirow{3}{*}{\rotatebox[origin=c]{90}{\parbox[c]{0.45cm}{\centering L.Club}}}
&Duchi
&56.0±5.5
&53.0±4.2
&52.7±2.9
&50.1±1.4
&50.1±0.9
&49.3±2.5
&65.7±0.2
\\ 
&Laplace
&50.1±0.6
&50.3±0.8
&49.6±1.1
&49.7±1.2
&49.9±0.7
&49.5±0.9
&65.7±0.2
\\\bottomrule
\end{tabular}
\end{sc}
\end{small}
\end{center}
\vskip -6mm
\end{table*}

%% file: tables/novel_class.tex
\begin{table*}[t]
\caption{Accuracy of classifiers for novel class classification, trained on data collected using different LDP mechanisms. Each column shows the $\epsilon$-LDP guarantee for the collected training set. Error bars represent $\pm 1$ standard deviation from the mean over 3 trials.}
\label{tab:hard_dist_shift}
\vspace{-4mm}
\begin{center}
\begin{small}
\begin{sc}
\begin{tabular}{p{1mm}p{22mm}|p{11mm}p{11mm}p{11mm}p{11mm}p{11mm}p{11mm}|p{13mm}}
\toprule

& \makecell[l]{Mechanism}
& \makecell[c]{$\eps=10$}
& \makecell[c]{$\eps=8$}
& \makecell[c]{$\eps=6$}
& \makecell[c]{$\eps=4$}
& \makecell[c]{$\eps=2$}
& \makecell[c]{$\eps=1$}
& \makecell[c]{No LDP}
\hspace{0.5cm}
\\
\midrule
&C-RLM (Ours)
&\textbf{89.3±0.5}
&\textbf{89.5±0.5}
&\textbf{89.7±0.6}
&\textbf{89.4±0.4}
&\textbf{87.2±1.4}
&\textbf{82.8±2.1}
&94.3±0.1
\\ 
\multirow{3}{*}{\rotatebox[origin=c]{90}{\parbox[c]{1cm}{\centering \,CIFAR10}}} 
&PrivUnit
&68.3±1.0
&66.7±2.0
&55.5±1.2
&52.1±0.6
&50.3±1.2
&50.3±2.0
&92.2±0.4
\\ 
&Duchi
&58.4±9.7
&57.8±9.4
&57.2±9.3
&56.9±8.9
&56.5±6.6
&52.6±2.8
&92.2±0.4
\\ 
&Laplace
&50.4±0.9
&50.2±0.6
&51.0±2.7
&49.1±0.6
&50.2±0.3
&49.8±0.4
&92.2±0.4
\\ \hline
&V-RLM (ours)
&\textbf{84.0±0.3}
&\textbf{80.9±1.8}
&\textbf{82.5±0.5}
&\textbf{80.7±0.3}
&\textbf{72.4±0.5}
&\textbf{55.8±8.7}
&94.0±0.4
\\ 
\multirow{3}{*}{\rotatebox[origin=c]{90}{\parbox[c]{1cm}{\centering \,\;MNIST}}}
&PrivUnit
&77.6±8.1
&71.0±2.6
&75.3±4.2
&54.5±3.9
&54.4±8.3
&54.6±10.1
&97.8±0.3
\\ 
&Duchi
&69.7±1.6
&68.7±3.1
&70.5±2.5
&66.5±6.5
&58.9±8.6
&50.4±0.2
&97.8±0.3
\\ 
&Laplace
&47.0±4.4
&47.3±4.0
&48.5±2.1
&48.7±2.0
&49.6±1.0
&48.5±2.2
&97.8±0.3
\\ \bottomrule
\end{tabular}
\end{sc}
\end{small}
\end{center}
\vskip -7mm
\end{table*}

%% file: tables/data_join.tex
\begin{table*}[t]
\caption{Accuracy of classifiers trained on the join of clean and $\epsilon$-LDP features of the Lending Club dataset. Each column shows the $\epsilon$-LDP guarantee for the collected training set. The baseline refers to the accuracy when classifying clean features only. Error bars represent $\pm 1$ standard deviation from the mean over 3 trials.}
\label{tab:data_join}
\vspace{-4mm}
\begin{center}
\begin{small}
\begin{sc}
\begin{tabular}{p{21mm}|p{9.5mm}p{9.5mm}p{9.5mm}p{9.5mm}p{9.5mm}p{10.5mm}|p{11.7mm}|p{13.9mm}}
\toprule
\makecell[l]{Mechanism}
& \makecell[c]{$\eps=10$}
& \makecell[c]{$\eps=8$}
& \makecell[c]{$\eps=6$}
& \makecell[c]{$\eps=4$}
& \makecell[c]{$\eps=2$}
& \makecell[c]{$\eps=1$}
& \makecell[c]{No LDP}
& \makecell[c]{Baseline}
\\
\midrule 
V-RLM (ours)
&\textbf{61.2±0.8}
&\textbf{60.5±0.4}
&\textbf{59.4±0.4}
&\textbf{57.9±0.7}
&\textbf{56.5±0.1}
&\textbf{56.1±0.1}
&64.8±0.2
&56.1±0.5
\\ 
Duchi
&56.5±0.2
&56.4±0.3
&56.2±0.3
&56.0±0.2
&55.9±0.2
&55.6±0.6
&65.8±0.6
&56.1±0.5
\\ 
Laplace
&56.8±0.5
&56.7±0.6
&56.7±0.2
&56.0±0.6
&55.8±0.3
&55.7±0.1
&65.8±0.6
&56.1±0.5
\\ \bottomrule
\end{tabular}
\end{sc}
\end{small}
\end{center}
\vskip -6.5mm
\end{table*}

%% file: tables/data_collection_private.tex
\begin{table*}[t]
\caption{Private Accuracy of classifiers trained on $\epsilon_\text{train}$-LDP (image, label) tuples collected using different LDP mechanisms. $\epsilon_\text{test}$ refers to the LDP guarantee of the images classified at inference time. Error bars represent $\pm 1$ standard deviation from the mean over 3 trials.}
\label{tab:data_collection_noisy}
\vspace{-4mm}
\begin{center}
\begin{small}
\begin{sc}
\begin{tabular}{p{1mm}|p{22mm}|p{15mm}p{11mm}p{11mm}p{11mm}p{11mm}p{11mm}|p{10mm}}
\toprule
&Mechanism
& $\epsilon_\text{train}=10$
& \makecell[c]{$8$}
& \makecell[c]{$6$}
& \makecell[c]{$4$}
& \makecell[c]{$2$}
& \makecell[c]{$1$}
& \makecell[c]{No}
\\
&
& $\,\epsilon_\text{test}=7.0$
& \makecell[c]{$5.6$}
& \makecell[c]{$4.2$}
& \makecell[c]{$2.8$}
& \makecell[c]{$1.4$}
& \makecell[c]{$0.7$}
& \makecell[c]{LDP}
\\
\midrule
&C-RLM (Ours)
&\textbf{45.0±0.3}
&\textbf{37.8±0.2}
&\textbf{27.8±0.5}
&\textbf{15.9±0.9}
&10.3±0.5
&\textbf{10.5±0.5}
&86.3±0.0
\\ 
\multirow{3}{*}{\rotatebox[origin=c]{90}{\parbox[c]{1cm}{\centering CIFAR10}}} 
&PrivUnit
&9.6±0.5
&10.0±0.4
&9.5±0.6
&10.1±0.7
&9.8±0.9
&10.4±0.8
&76.1±0.7
\\ 
&Duchi
&10.5±1.2
&10.0±0.7
&9.8±0.6
&9.9±0.7
&9.6±0.5
&10.3±0.5
&76.1±0.7
\\ 
&Laplace
&10.1±0.5
&10.2±1.3
&9.7±0.3
&10.4±0.8
&\textbf{10.6±0.6}
&9.8±0.3
&76.1±0.7
\\ \hline
&V-RLM (ours)
&\textbf{42.3±0.5}
&\textbf{37.7±0.5}
&\textbf{31.7±1.2}
&\textbf{20.1±0.6}
&\textbf{10.5±0.6}
&10.2±0.5
&94.9±0.2
\\ 
\multirow{3}{*}{\rotatebox[origin=c]{90}{\parbox[c]{1cm}{\centering \;\;MNIST}}}
&PrivUnit
&9.6±0.3
&10.0±1.3
&10.3±0.4
&11.0±0.7
&10.3±0.2
&\textbf{10.8±1.2}
&96.0±0.4
\\ 
&Duchi
&10.1±0.4
&10.1±0.3
&10.4±0.1
&10.1±0.9
&9.7±0.9
&10.2±0.7
&96.0±0.4
\\ 
&Laplace
&10.1±0.3
&10.7±0.8
&11.0±0.5
&9.8±0.9
&9.9±0.3
&10.6±1.3
&96.0±0.4
\\ \hline
&Maximum
&80.7
&69.3
&53.8
&36.0
&20.0
&14.2
&100.0
\\ \bottomrule
\end{tabular}
\end{sc}
\end{small}
\end{center}
\vskip -6mm
\end{table*}

%% file: tables/hparams_datajoin.tex
\begin{table}[ht]
\caption{V-RLM hyperparameters used for data join experiments on the Lending Club dataset.}
\label{Table:hyperparameters_data_join}
\begin{center}
\begin{small}
\begin{sc}
\begin{tabular}{p{20mm}p{15mm}p{15mm}p{15mm}}
\toprule
$\epsilon$
&$d$
&$l$
&$\epsilon_\text{pre-training}$
\\
\midrule
$\infty$  &8  &5 &20\\
10 &8  &5 &10 \\
8  &5  &5 &15\\
6  &5  &5 &15\\
4  &5  &5 &10\\
2  &5  &5 &15\\
1  &5  &5 &10\\
\bottomrule
\end{tabular}
\end{sc}
\end{small}
\end{center}
\vskip -0.1in
\end{table}

%% file: tables/hparams_dpadam.tex
\begin{table}[ht]
\caption{DP-Adam hyperparameters used for the V-RLM data collection and novel-class classification under CDP.}
\label{Table:dp_hyperparameters}
\begin{center}
\begin{small}
\begin{sc}
\begin{tabular}{p{11mm}|p{10mm}p{11mm}p{10mm}p{17mm}}
\toprule
Task
&$\epsilon_\text{CDP}$
&Learning Rate
&Batch Size
&Noise Multiplier
\\
\midrule
MNIST              &5    &5e-4  &64  &0.7  
\\
                   &1    &5e-4  &64  &1.1  
\\ \midrule
Lending            &5   &1e-4  &128  &0.56  
\\
Club               &1   &1e-4  &128  &1.1  
\\ \bottomrule
\end{tabular}
\end{sc}
\end{small}
\end{center}
\vskip -0.1in
\end{table}

%% file: tables/hparams_mechanisms.tex
\begin{table*}[t]
\caption{V-RLM and C-RLM hyperparameters used for the data collection and novel-class classification experiments.}
\label{Table:hyperparameters_data_collection}
\begin{center}
\begin{small}
\begin{sc}
\begin{tabular}{p{.2\textwidth} | p{.14\textwidth} p{.08\textwidth} p{.08\textwidth} p{.08\textwidth} p{.08\textwidth} p{.08\textwidth}}
\toprule
Experiment &Task &$\epsilon$ &$\lambda$ &$d$ &$l$ &$\epsilon_\text{pre-training}$
\\ \midrule
                 &CIFAR-10    &$\infty$ &N/A  &32 &5  &No noise
\\
                   &                &10  &0.7 &32  &5  &No noise
\\
                   &                &8   &0.7 &32  &5 &No noise
\\
                   &                &6   &0.7 &32  &5 &70
\\
                   &                &4   &0.7 &32  &5  &30
\\
                   &                &2   &0.7 &32  &5  &20
\\
                   &                &1   &0.7 &32  &5 &20
\\                   
 &MNIST           &$\infty$ &N/A &8 &10  &Learnt
\\
&                &10  &0.7 &8 &10  &33
\\
Classifying clean  &                &8   &0.7 &8 &5  &32 
\\
datapoints (Section  &                &6   &0.7 &8 &5  &19 
\\
\ref{Section:experiments_clean} experiments)   &                &4   &0.7 &8 &7.5  &13 
\\
                   &                &2   &0.7 &8 &7.5  &7
\\
                   &                &1   &0.7 &8 &5  &7
\\
  &Lending Club    &$\infty$ &N/A &8 &10  &Learnt
\\
  &                &10  &0.7 &8 &5  &15
\\
  &                &8   &0.7 &8 &5 &29
\\
                   &                &6   &0.7 &8 &5 &29
\\
                   &                &4   &0.7 &8 &5  &15
\\
                   &                &2   &0.95 &8 &5  &15
\\
                   &                &1   &0.95 &8 &10 &21
\\ \hline
&CIFAR-10           &$\infty$ &N/A &32 &5 &No noise
\\
&                &10  &0.7 &32 &5  &27
\\
&                &8   &0.7 &32 &5  &24
\\
                   &                &6   &0.7 &32 &5  &20 
\\
                   &                &4   &0.7 &32 &5 &15 
\\
                   &                &2   &0.7 &32 &5 &8
\\
Classifying LDP   &                &1   &0.7 &32 &5  &8
\\
datapoints (Section &MNIST           &$\infty$ &N/A &8 &10  &Learnt
\\
\ref{Section:experiments_private} experiments)&                &10  &0.7 &8 &10  &5
\\
                   &                &8   &0.7 &8 &7.5  &5
\\
                   &                &6   &0.7 &8 &7.5  &5 
\\
                   &                &4   &0.7 &8 &5 &5 
\\
                   &                &2   &0.7 &8 &5 &15
\\
                   &                &1   &0.7 &8 &7.5  &15
\end{tabular}
\end{sc}
\end{small}
\end{center}
\end{table*}

%% file: tables/data_collection_cdp.tex
\begin{table*}[t]
\caption{Accuracy of classifiers trained on data collected using different LDP mechanisms, trained using DP-Adam. Each column shows the $\epsilon$-LDP guarantee for the collected training set. Error bars represent $\pm 1$ standard deviation from the mean over 3 trials.}
\label{tab:cdp_data_collection}
\begin{center}
\begin{small}
\begin{sc}
\begin{tabular}{p{1mm}p{27mm}|p{11mm}p{11mm}p{11mm}p{10mm}p{10mm}p{11mm}|p{10mm}}
\toprule

& \makecell[l]{Mechanism}
& \makecell[c]{$\eps=10$}
& \makecell[c]{$\eps=8$}
& \makecell[c]{$\eps=6$}
& \makecell[c]{$\eps=4$}
& \makecell[c]{$\eps=2$}
& \makecell[c]{$\eps=1$}
& \makecell[c]{No\\ LDP}
\hspace{0.5cm}
\\
\midrule
&V-RLM ($\eps_\text{cdp}=\infty$)
&\textbf{86.1±1.0}
&\textbf{82.1±1.6}
&\textbf{72.8±3.2}
&\textbf{61.3±2.8}
&\textbf{35.3±9.7}
&16.9±1.3
&94.9±0.2
\\ 
\multirow{3}{*}{\rotatebox[origin=c]{90}{\parbox[c]{1cm}{\centering MNIST}}}
&V-RLM ($\eps_\text{cdp}=5$)
&78.6±1.2
&75.9±0.9
&66.2±2.4
&50.7±2.5
&16.8±4.4
&\textbf{17.2±1.3}
&87.2±0.4
\\ 
&V-RLM ($\eps_\text{cdp}=1$)
&72.6±1.4
&70.3±1.9
&60.3±2.6
&46.9±3.3
&17.1±3.2
&14.4±4.3
&83.7±0.5
\\ 
&PrivUnit
&38.2±4.6
&15.1±4.2
&12.4±1.6
&9.2±2.6
&9.5±2.9
&11.3±4.5
&96.0±0.4
\\ 
&Duchi
&13.9±4.5
&14.3±5.0
&13.3±3.8
&14.1±4.9
&14.0±5.6
&10.0±1.1
&96.0±0.4
\\ 
&Laplace
&\;\;9.2±1.4
&\;\;9.8±2.8
&10.6±0.3
&10.0±0.6
&\;\;9.0±1.1
&10.0±0.8
&96.0±0.4
\\ \hline
\multirow{3}{*}{\rotatebox[origin=c]{90}{\parbox[c]{2cm}{\centering L. Club}}}
&V-RLM ($\eps_\text{cdp}=\infty$)
&\textbf{63.7±0.6}
&63.2±0.3
&\textbf{62.6±0.5}
&61.1±1.2
&\textbf{53.9±4.3}
&\textbf{55.1±4.5}
&65.0±0.3
\\ 
&V-RLM ($\eps_\text{cdp}=5$)
&\textbf{63.7±0.4}
&\textbf{63.4±0.2}
&\textbf{62.6±0.5}
&61.3±1.4
&53.6±4.1
&54.9±4.5
&65.1±0.3
\\ 
&V-RLM ($\eps_\text{cdp}=1$)
&63.2±0.5
&63.1±0.4
&\textbf{62.6±0.6}
&\textbf{61.5±1.5}
&\textbf{53.9±2.4}
&54.6±4.1
&64.6±0.2
\\ 
&Duchi
&56.0±5.5
&53.0±4.2
&52.7±2.9
&50.1±1.4
&50.1±0.9
&49.3±2.5
&65.7±0.2
\\ 
&Laplace
&50.1±0.6
&50.3±0.8
&49.6±1.1
&49.7±1.2
&49.9±0.7
&49.5±0.9
&65.7±0.2
\\ \bottomrule
\end{tabular}
\end{sc}
\end{small}
\end{center}
\vskip -0.1in
\end{table*}

%% file: tables/ablation_datasplit.tex
\begin{table*}[t]

\caption{Accuracy of classifiers trained on data collected using different LDP mechanisms. $\eta$ represents the proportion of MNIST dataset used for V-RLM training. For the benchmarks, the entire MNIST dataset was used for classifier training. Each column shows the $\epsilon$-LDP guarantee for the collected training set. Error bars represent $\pm 1$ standard deviation from the mean over 3 trials.}
\label{tab:ablation_datasplit}
\begin{center}
\begin{small}
\begin{sc}
\begin{tabular}{p{23mm}p{11mm}|p{12mm}p{12mm}p{12mm}p{12mm}p{12mm}p{12mm}}
\toprule
\makecell[l]{Mechanism}
& $\eta$
& \makecell[c]{$\eps=10$}
& \makecell[c]{$\eps=8$}
& \makecell[c]{$\eps=6$}
& \makecell[c]{$\eps=4$}
& \makecell[c]{$\eps=2$}
& \makecell[c]{$\eps=1$}
\hspace{0.5cm}
\\
\midrule
&0.75
&86.1±1.0
&82.1±1.6
&72.8±3.2
&61.3±2.8
&35.3±9.7
&16.9±1.3
\\ 
V-RLM
&0.50
&\textbf{86.4±0.2}
&\textbf{84.0±1.0}
&\textbf{78.0±3.1}
&\textbf{61.6±2.3}
&35.2±1.7
&14.1±1.9
\\ 
&0.25
&85.1±0.2
&81.9±0.5
&75.4±2.7
&60.5±3.5
&\textbf{44.8±4.0}
&16.3±0.8
\\
&0.10
&81.1±1.8
&77.4±1.7
&67.9±1.4
&57.6±3.5
&37.9±1.4
&\textbf{18.0±2.7}
\\
\midrule
PrivUnit
&-
&47.3±2.3
&29.2±5.5
&16.2±2.7
&12.0±3.7
&10.7±0.7
&14.3±0.4
\\
Duchi
&-
&26.3±3.0
&20.5±0.8
&20.4±2.9
&17.0±2.7
&14.0±2.7
&13.9±1.0
\\
Laplace
&-
&16.6±0.8
&16.5±4.2
&11.8±1.3
&12.1±1.6
&11.1±1.2
&11.5±0.6
\\ \bottomrule
\end{tabular}
\end{sc}
\end{small}
\end{center}
\vskip -0.1in
\end{table*}

%% file: tables/data_collection_pca.tex
\begin{table*}[t]
\caption{Accuracy of classifiers trained on LDP data, collected using different dimensionality reduction techniques. Each column shows the $\epsilon$-LDP guarantee for the collected training set. Error bars represent $\pm 1$ standard deviation from the mean over 3 trials.}
\label{tab:data_collection_pca}
\begin{center}
\begin{small}
\begin{sc}
\begin{tabular}{p{1mm}p{22mm}|p{11mm}p{11mm}p{11mm}p{11mm}p{11mm}p{11mm}|p{13mm}}
\toprule

& \makecell[l]{Mechanism}
& \makecell[c]{$\eps=10$}
& \makecell[c]{$\eps=8$}
& \makecell[c]{$\eps=6$}
& \makecell[c]{$\eps=4$}
& \makecell[c]{$\eps=2$}
& \makecell[c]{$\eps=1$}
& \makecell[c]{No LDP}
\hspace{0.5cm}
\\
\midrule
&V-RLM
&\textbf{86.1±1.0}
&\textbf{82.1±1.6}
&\textbf{72.8±3.2}
&\textbf{61.3±2.8}
&\textbf{35.3±9.7}
&\textbf{16.9±1.3}
&94.9±0.2
\\ 
\multirow{3}{*}{\rotatebox[origin=c]{90}{\parbox[c]{1cm}{\centering MNIST}}}
&PCA
&17.3±8.1
&15.1±4.9
&15.0±5.4
&14.8±4.9
&15.1±5.6
&15.7±6.6
&86.4±0.3
\\ 
&PrivUnit
&38.2±4.6
&15.1±4.2
&12.4±1.6
&9.2±2.6
&9.5±2.9
&11.3±4.5
&96.0±0.4
\\ 
&Duchi
&13.9±4.5
&14.3±5.0
&13.3±3.8
&14.1±4.9
&14.0±5.6
&10.0±1.1
&96.0±0.4
\\ 
&Laplace
&\;\;9.2±1.4
&\;\;9.8±2.8
&10.6±0.3
&10.0±0.6
&\;\;9.0±1.1
&10.0±0.8
&96.0±0.4
\\ \bottomrule
\end{tabular}
\end{sc}
\end{small}
\end{center}
\vskip -0.1in
\end{table*}

%% file: tables/fig_proof_equidistant.tex
\begin{figure}[ht]
\begin{center}
\begin{subfigure}{ }
  \begin{tikzpicture}[domain=-2.5:4] 
    \draw[very thin,color=gray] (-2.6,-2.6) grid (4.2,4.2);
    \draw[<->] (-2.7,0) -- (4.2,0) node[right] {$x_1$}; 
    \draw[<->] (0,-2.7) -- (0,4.2) node[above] {$x_2$};
    \draw [fill=red, red, fill opacity=.5, opacity=.5] (2,2) rectangle (4.3,4.3);
    \draw [fill=red, red, fill opacity=.5, opacity=.5] (-2.6,-2.6) rectangle (0,0);
    \draw[color=red, domain=0:2]    plot (\x,\x);
    \draw[color=red, domain=2:4]    plot (\x,2);
    \draw[color=red, domain=-2.5:0]    plot (\x,0);
    \draw[color=red, domain=2:4]    plot (2,\x);
    \draw[color=red, domain=-2.5:0]    plot (0,\x);
    \filldraw (0,2) circle (0.08cm) node (c2) {} node[anchor=east,fill=none,yshift=0.2cm] {$c^{(2)}$};
    \filldraw (2,0) circle (0.08cm) node (c1) {} node[anchor=east,fill=none,yshift=-0.2cm] {$c^{(1)}$};
  \end{tikzpicture}
\caption{Red shaded areas and lines represent the regions of $\mathbb{R}^2$ in which all points are equidistant from $c^{(1)}$ and $c^{(2)}$.}
\label{vlm_fig:equidistant_left}
\end{subfigure}%
\vspace*{\fill}   
\begin{subfigure}{}
  \begin{tikzpicture}[domain=-2.5:4] 
    \draw[very thin,color=gray] (-2.6,-2.6) grid (4.2,4.2);
    \draw[<->] (-2.7,0) -- (4.2,0) node[right] {$x_1$}; 
    \draw[<->] (0,-2.7) -- (0,4.2) node[above] {$x_2$};
    \draw[color=red, domain=-2.7:4.2]    plot (\x,\x) node[anchor=east,fill=none,yshift=-0.0cm, xshift=-0.15cm, rotate=45] {\scriptsize Decision boundary}; ;
    \filldraw (0,2) circle (0.08cm) node (c2) {} node[anchor=east,fill=none,yshift=0.2cm] {$c^{(2)}$};
    \filldraw (2,0) circle (0.08cm) node (c1) {} node[anchor=east,fill=none,yshift=-0.2cm] {$c^{(1)}$};
  \end{tikzpicture}
\caption{The red line represents the decision boundary that separates $c^{(1)}$ and $c^{(2)}$ equally in L1 distance. Regions in which points are equidistant from representations $c^{(1)}$ and $c^{(2)}$ are divided based on the closest representation in L2 distance.}
\label{vlm_fig:equidistant_right}
\end{subfigure}%
\label{vlm_fig:equidistant}
\end{center}
\end{figure}

%% file: tables/fig_proof_boundary1.tex
\begin{figure}[ht]
\begin{center}
  \begin{tikzpicture}[domain=-2.5:2.5] 
    \draw[very thin,color=gray] (-2.6,-2.6) grid (2.6,2.6);
    \draw[<->] (-2.8,0) -- (2.8,0) node[right] {$x_1$}; 
    \draw[<->] (0,-2.8) -- (0,2.8) node[above] {$x_2$};
    \draw[color=red]    plot (\x,\x) node[anchor=east,fill=none,yshift=-0.0cm, xshift=-0.15cm, rotate=45] {\scriptsize Decision boundary}; 
    \draw[color=red]    plot (\x,-\x);
    \filldraw (0,2) circle (0.08cm) node (c2) {} node[anchor=east,fill=none,yshift=0.2cm] {$c^{(2)}$};
    \filldraw (2,0) circle (0.08cm) node (c1) {} node[anchor=east,fill=none,yshift=-0.2cm] {$c^{(1)}$};
    \filldraw (0,-2) circle (0.08cm) node (c4) {} node[anchor=east,fill=none,yshift=-0.2cm] {$c^{(4)}$};
    \filldraw (-2,0) circle (0.08cm) node (c3) {} node[anchor=east,fill=none,yshift=-0.2cm] {$c^{(3)}$};
    \draw[dotted, color=blue, line width=0.5mm, opacity=0.8, domain=0:2]    plot (\x,2-\x);
    \draw[dotted, color=blue, line width=0.5mm, opacity=0.8, domain=-2:0]    plot (\x,-2-\x);
    \draw[dotted, color=blue, line width=0.5mm, opacity=0.8, domain=0:2]    plot (\x,\x-2);
    \draw[dotted, color=blue, line width=0.5mm, opacity=0.8, domain=-2:0]    plot (\x,\x+2) node[anchor=east,fill=none,yshift=-1.7cm, xshift=-1.0cm] {$\mathcal{Z}$};
  \end{tikzpicture}
\end{center}
\caption{The decision boundary for a classifier that equally separates (in $\ell_1$-distance) vertices $c^{(i)}$ for $i \in \{1,2,3,4\}$ in 2-dimensional space. The blue region denotes the taxicab sphere $\mathcal{Z}$.}
\label{vlm_fig:decision_boundary}
\end{figure}
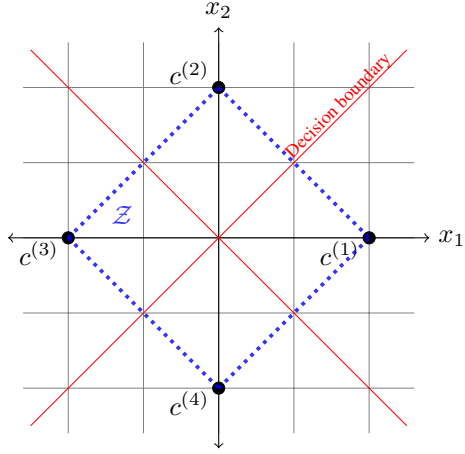